\pdfoutput=1
\documentclass{article}
\usepackage{iclr/format_iclr/iclr2027_conference,times}
\iclrfinalcopy

\usepackage{graphicx}
\usepackage{booktabs}
\usepackage{float}
\usepackage{placeins}
\usepackage{multirow}
\usepackage{amsfonts}
\usepackage{mathtools}
\usepackage{siunitx}
\usepackage{makecell}
\usepackage{adjustbox}
\usepackage{tabularx}
\usepackage{lscape}
\makeatletter
\g@addto@macro\landscape{\global\pdfpageattr{/Rotate 90}}
\g@addto@macro\endlandscape{\global\pdfpageattr{/Rotate 0}}
\newenvironment{widetables}
  {\begin{landscape}\renewcommand\table[1][]{\@float{table}[H]}}
  {\end{landscape}}
\makeatother
\usepackage{afterpage}
\usepackage{subcaption}
\usepackage{algorithm}
\usepackage{algpseudocode}
\usepackage{xspace}
\usepackage{enumitem}
\usepackage{tikz}
\usepackage{hyperref}
\usetikzlibrary{positioning, arrows.meta, calc}

\usepackage{xcolor}

\newif\ifrevisionfinal
\revisionfinalfalse
\ifrevisionfinal
  
  \newcommand{\OLD}[1]{}
\else
  
  \newcommand{\OLD}[1]{{\color{red}#1}}
\fi

\newcommand{\Scarce}[1]{\ensuremath{\mathrm{Scarce}_{#1}}}

\graphicspath{{figures/}{paper/figures/}}
\usepackage{amsthm}
\theoremstyle{plain}

\theoremstyle{definition}

\title{On the Effectiveness of Resampling for Supervised Graph Anomaly Detection}

\author{Cyril Perosino\textsuperscript{1}, Rino Di Chiara\textsuperscript{2} \& Hamida Seba\textsuperscript{1}\\
\textsuperscript{1}Universit\'e Claude Bernard Lyon 1\\
\textsuperscript{2}Politecnico di Torino\\
\texttt{cyril.perosino@univ-lyon1.fr}}

\hypersetup{
  pdftitle={On the Effectiveness of Resampling for Supervised Graph Anomaly Detection},
  pdfauthor={Cyril Perosino, Rino Di Chiara, Hamida Seba},
  pdfkeywords={graph anomaly detection, resampling, oversampling, class imbalance, benchmark}
}

\begin{document}

\maketitle
\lhead{Preprint}

\begin{abstract}   
Class imbalance and scarce anomaly labels motivate oversampling in
supervised graph anomaly detection (GAD), but its benefits across
detectors and benchmarks remain unclear. In the low-label setting of
GADBench, with 20 labelled anomalies and 80 labelled normal nodes, we
compare class weighting and up to six oversampling strategies in ten
detector configurations from three families: graph neural networks
(GNNs), tree ensembles on neighbourhood-aggregated features, and tabular
foundation models, on 17 datasets. On the eight datasets with organic
anomalies, no strategy has a statistically resolved average gain in any
configuration, and which nodes are labelled matters as much as the
strategy. On the nine datasets with injected anomalies, Gaussian-noise
oversampling (ROSE) gains up to 5.2 AUPRC points in four of the six tree
and tabular foundation model configurations, and in all six only on
attribute-swap anomalies; in four GNNs its gain is not distinguishable
from zero. This gain depends on the injection recipe, which gives swapped
nodes an atypical feature norm: it is no longer distinguishable from zero
when the swap uses a single random donor or preserves the norm, although
most detectors then also stop detecting the swaps. Over the 17 datasets,
tuning the tree detectors adds about three points, as much as ROSE, and
cuts ROSE's gain by about half or more. With the same source nodes and
class weight, sampled graph views are not distinguishable from plain
copies of those nodes. We recommend tuning baselines, repeating labelled
splits, reporting organic and injected anomalies separately, and checking
label-free degree and feature-norm baselines.
\end{abstract}

\section{Introduction}
\label{sec:introduction}

Graph anomaly detection (GAD) identifies unusual nodes, edges or substructures
in applications such as fraud detection and cybersecurity.
We study supervised node anomaly detection with few labels.
Labelled anomalies are scarce in fraud, abuse and intrusion graphs:
confirming a case is costly, and the standard benchmark for this setting
trains on twenty labelled anomalies and eighty labelled normal nodes
\citep{tang2023gadbenchrevisitingbenchmarkingsupervised}. With so few
positives, the cheapest lever is the training set itself: reweight it,
duplicate it, perturb it, or generate new graph observations. Many such
methods exist, from random oversampling to diffusion-based graph
generators \citep{chawla2002smote,rose,graphsmote,godm}, and each is
usually reported with its own detector, representation and split. A
practitioner cannot tell whether a gain comes from the new
samples, from which observations were selected, from the extra class
weight or from the detector.

Two notions recur throughout. \emph{Resampling} changes the labelled
training set before a detector is fitted and leaves the detector
unchanged: class weighting gives the rare class more weight, and
oversampling adds rows to it, either copies of labelled anomalies or
synthetic rows built from them. It is the standard remedy for class
imbalance on tabular data, where its benefit is itself debated
\citep{sakho2026rebalancing}. Benchmarks, for their part, are of two
kinds. In some the anomalies come with the data, for example accounts
flagged as fraudulent; we call them \emph{organic}. In others anomalies
are \emph{injected} into a graph that has none, by replacing the
attributes of some nodes or by wiring small groups of nodes into cliques.
The two kinds need not respond to resampling in the same way, and we
report them separately.

Two problems arise from the literature. First, a
sampler acts on a representation, and the three major detector families in use
today build theirs very differently: a graph neural network (GNN) learns
it, a tree ensemble receives fixed neighbourhood aggregates, and a tabular
foundation model reads the labelled rows in context built from the graph. 
A result validated in
one family may not be true in another. Second, organic and synthetic
graphs have different sensitivity to resampling.
Synthetic graph evaluation generally relies on anomalies injected by a standard recipe
\citep{dominant,cola}, which is known to leave label-free traces, a
high degree and a large feature norm, that alone match unsupervised
detectors \citep{huang2023unsupervised}. Whether
these traces drive the gains of supervised resampling has not been
examined.

We ask \emph{when resampling improves supervised node anomaly detection},
and answer with a controlled empirical study that promotes no method. 
Our organizing lens is \textbf{allocation, transformation and mass}
($\mathcal A/\mathcal T/\mathcal M$, Section~\ref{sec:atm}): which
observations receive the resampling budget, how their representations
change, and how much training weight their class receives.

We compare
class weighting and up to six resampling strategies, plus a seventh added
last (MGS; \citealp{sakho2026rebalancing}), in ten detector
configurations on 17 benchmarks, with ten paired labelled splits and two
references (unweighted training and class weighting). The configurations
cover the three major detector families: TabPFN-v2 on two tables, one of them the
recent TFM4GAD \citep{tfm4gad,tabpfnv2}; the two tree detectors of
GADBench, one of which (XGBGraph) recent GNN papers keep as a baseline
\citep{spacegnn,apf,sagad,chigad}; and four GNNs,
including GHRN and SpaceGNN \citep{ghrn,spacegnn}. No family can be left
out: on our splits, with no model tuned, the tree detector RFGraph has a
higher mean AUPRC than each GNN over the 17 benchmarks (33.2 against at
most 29.9) and TFM4GAD is first (39.0).

Our findings (Table~\ref{tab:summary} pairs each with its main evidence
and its main limit):
\begin{enumerate}[leftmargin=*,label=\textbf{\arabic*.},itemsep=1pt,topsep=2pt]
\item \textbf{No statistically resolved average gain on eight
organic-anomaly benchmarks.} None
of the six samplers has a statistically resolved panel-level gain in any
configuration, and for ROSE the intervals exclude gains above two points in eight
of the ten configurations. Single benchmarks move in both directions,
choosing the strategy on validation data gains little
(Section~\ref{sec:rq1}), and no organic gain appears when the label
count and the class ratio are varied separately
(Section~\ref{sec:budget}).
\item \textbf{On injected benchmarks, noise helps tabular detectors,
consistently only on attribute swaps.} ROSE gains 2.6 to 10.8 points on
attribute-swap anomalies in all six tabular configurations. In four GNNs
it has no gain, at the hidden layer or at the input
(Section~\ref{sec:injected}).
\item \textbf{The swap gain depends on the injection recipe.}
The standard recipe copies the most distant of 50 candidates, which
shifts the feature norm of swapped nodes, upwards on most graphs. With
fewer candidates the norm signal, the detectability of the swaps and
ROSE's gain shrink together; with a random donor or a norm-preserving
swap no gain is resolved, and most detectors no longer detect the swaps
(Section~\ref{sec:leak}).
\item \textbf{Tuning a tree detector is worth as much as ROSE.} Over
the 17 benchmarks a validation search adds about three points to each
tree detector and cuts ROSE's gain by about half or more; only XGBGraph's
remains resolved (Section~\ref{sec:tuning}).
\item \textbf{Graph views show no detectable benefit over copies within
either panel.} With source selection and class weight matched, views of a
node's sampled neighbourhood are not distinguishable from copies of the node on
the organic or the injected panel, for two tree learners; over all 17
benchmarks, one of two RFGraph variants of the control gives a small gain
(Section~\ref{sec:rq2}).
\end{enumerate}

\begin{table}[t]
\centering\small
\setlength{\tabcolsep}{4pt}
\caption{\textbf{Summary of the findings}, each with its main evidence
and its main limit. Gains are in AUPRC points; \emph{resolved}: the 95\%
interval excludes zero.}
\label{tab:summary}
\begin{tabularx}{\linewidth}{@{}>{\raggedright\arraybackslash}p{3.2cm}>{\raggedright\arraybackslash}X>{\raggedright\arraybackslash}X@{}}\toprule
Finding & Main evidence & Main limit\\\midrule
1. No resolved average gain on the organic panel
& No sampler has an interval above zero in any of the ten configurations;
ROSE's interval excludes gains above two points in eight of them
(Section~\ref{sec:rq1}, Table~\ref{tab:cross-family})
& Eight benchmarks: an interval that contains zero does not exclude a
meaningful gain, and single benchmarks move in both directions\\
\addlinespace[3pt]
2. On the injected panel, noise helps tabular detectors, consistently
only on attribute swaps
& ROSE gains $+2.5$ to $+5.2$, resolved in four of the six tabular
configurations; on swaps, $+2.6$ to $+10.8$, resolved in all six; no gain
in four GNNs, at the hidden layer or at the input
(Section~\ref{sec:injected}, Figure~\ref{fig:rose})
& The swap gain is carried by five of eight benchmarks; the GNNs are
untuned\\
\addlinespace[3pt]
3. The swap gain depends on the injection recipe
& On nine graphs that we inject ourselves, the gain grows with the norm
leak along a candidate ladder and is not resolved with a random donor or
a norm-preserving swap (Section~\ref{sec:leak}, Table~\ref{tab:recipe})
& Removing the leak also makes the swaps hard to detect; the tests do not
separate the two\\
\addlinespace[3pt]
4. Tuning a tree detector is worth as much as ROSE
& Over the 17 benchmarks a validation search adds about three points to
each tree detector; ROSE's gain falls from $+3.2$ to $+1.8$ (XGBGraph)
and from $+2.8$ to $+0.6$ (RFGraph) (Section~\ref{sec:tuning})
& The search also selects the aggregation and uses validation labels that
the default trees never see; only the trees were tuned\\
\addlinespace[3pt]
5. Graph views show no detectable benefit over copies
& With sources and class mass matched, views minus copies is unresolved
in both panels for both tree learners (Section~\ref{sec:rq2},
Table~\ref{tab:matched})
& Trees only, and no positive control on the organic panel; one of two
RFGraph variants gives $+0.44$ over all 17 benchmarks\\
\bottomrule
\end{tabularx}
\end{table}

Code and per-trial results:
\url{https://github.com/Peppino58/GraphResampling4GAD}
(tag \texttt{arxiv-v1} for this version).

\section{Related Work}
\label{sec:related}

\paragraph{Supervised graph anomaly detection.}
Recent detectors are mostly GNNs: spectral and heterophily-aware models
such as BWGNN \citep{bwgnn}, GHRN \citep{ghrn} and PMP \citep{pmp};
sampling-based PC-GNN \citep{pcgnn}; and, for limited supervision,
consistency training \citep{consisgad}, multi-space ensembles
\citep{spacegnn} and pre-training \citep{apf}. GADBench
\citep{tang2023gadbenchrevisitingbenchmarkingsupervised} showed that tree
ensembles on neighbourhood-aggregated features exceed, on average and at
default hyperparameters, the GNNs available in 2023 when trained on
twenty labelled anomalies, and XGBGraph remains a baseline of recent
detectors \citep{spacegnn,apf,sagad,chigad}. The family also includes
LightGAD \citep{lightgad}, on a homophily-weighted aggregation, and
balanced forests \citep{balancedrf}; we keep GADBench's two detectors. TFM4GAD \citep{tfm4gad}
applies a tabular foundation model \citep{tabpfnv2} to a flattened graph
table and reports, on four GADBench datasets, a higher mean AUPRC than
eight trained baselines, including SpaceGNN, ConsisGAD and both tree
detectors. For node property prediction at large, boosted trees on
neighbourhood aggregates can be strong baselines, although GNNs outperform
them on most datasets \citep{graphland}; see also
\citet{grinsztajn2022tree,adbench} for tabular data.

\paragraph{Resampling under class imbalance.}
Tabular resampling spans duplication (random oversampling, ROS), kernel
perturbation \citep[ROSE,][]{rose} and interpolation
\citep[SMOTE and ADASYN,][]{chawla2002smote,adasyn}. IGL-Bench \citep{iglbench} compares imbalance methods across GNN
backbones for node and graph classification, and BAT \citep{bat} reduces
class-imbalance bias there by topological augmentation, without
rebalancing; neither covers anomaly detection, tabular detectors or
injected benchmarks. With tuned random forests, gradient boosting or
logistic regression, applying no rebalancing is competitive on most
tabular datasets, less so under extreme imbalance
\citep{sakho2026rebalancing}. The same study proposes Multivariate
Gaussian SMOTE (MGS), which draws each synthetic point from a Gaussian
fitted to a minority point and its $K$ nearest minority neighbours, and
reports default random forests on par with tuned ones with it. An MGS
draw stays in the affine span of those points, whereas ROSE perturbs
every feature independently; we run MGS for that contrast
(Section~\ref{sec:injected}).
\citet{ma2026augmentation} find that, for a
well-specified score model, synthetic augmentation brings no
population-level gain in ranking metrics beyond a possible finite-sample
variance reduction, and that systematic gains require misspecification.

\paragraph{Injected anomalies.}
The standard recipe \citep{dominant,cola} creates two types of anomaly:
attribute swaps and cliques (Section~\ref{sec:framework}).
\citet{huang2023unsupervised} traced a leak in both, to the size of the
candidate set and the Euclidean distance of the swap and to the degree of
clique members, showed that a degree-and-norm score rivals unsupervised
detectors, recommended comparing against it, and suggested recipes with
fewer candidates, another distance or degree-preserving edges, on which
our variants in Section~\ref{sec:leak} build;
\citet{liu2022bondbenchmarkingunsupervisedoutlier} report a wide gap
between synthetic and organic outliers, and \citet{gu2023three} propose
more diverse injections. These studies concern unsupervised detection.
We study the supervised case, where the leak is not exploited
automatically: with 20 labelled anomalies, default trees and TabPFN score the swaps
far below what the norm alone achieves, and a sampler's gain appears
where that signal is present.

\paragraph{Graph-aware and adaptive resampling.}
Graph-aware methods use the graph when resampling. Some generate new
observations: GraphSMOTE \citep{graphsmote}, GODM \citep{godm} and GenGA
\citep{genga} generate nodes with edges, DAGAD \citep{dagad} augments
learned node representations, and GraphENS \citep{graphens} synthesises
ego-networks. Others keep the nodes and change their graph context, as
subgraph samplers such as Forest Fire Sampling
\citep{forestfiresampling}, ShaDow \citep{shaDow} and GRAPES
\citep{younesian2025grapeslearningsamplegraphs} do. We evaluate one
method of each kind, GODM and forest-fire island views
(Section~\ref{sec:strategies}). TAM \citep{tam}
and ReNode \citep{renode} change the objective instead of the data and
lie outside the operators evaluated here.
Resampling also differs in \emph{allocation}, that is, in which
observations receive the budget: ADASYN \citep{adasyn} gives more
synthetic samples to minority points that are locally difficult, GOSS
\citep{gosslightgbm} keeps the rows with large gradients, and GraphSHA
\citep{graphsha} synthesises harder minority samples. Such methods change
the selection of the sources and their transformation together, and often
the total weight of a class as well, so a gain cannot be attributed to
one of the three. Our interface separates them
(Section~\ref{sec:atm}), and matched controls vary one at a time
(Section~\ref{sec:adaptive-resampling}).

\section{Study Design}
\label{sec:framework}

We study transductive node anomaly detection. Following GADBench, each
method receives 20 labelled anomalies and 80 labelled normal nodes, with
a disjoint validation set of the same size; we write \Scarce{20} for this
budget; Section~\ref{sec:budget} varies the number of labels and the
class ratio separately. We evaluate on 17 datasets
(Appendix~\ref{sec:datasets}). Eight have \emph{organic} anomalies,
labelled in the original data: Amazon, FDCompCN, Questions, Reddit,
T-Finance, Tolokers, Weibo and YelpChi. Nine have \emph{injected}
anomalies, planted in an anomaly-free graph by the standard recipe:
ACM, BlogCatalog, Citeseer, Cora, Flickr, Pubmed, Amazon-Computers,
Amazon-Photo and Coauthor-CS. We call the two groups the organic and the
injected panel. The recipe creates two types in equal or
near-equal numbers: \emph{attribute-swap} anomalies, whose attributes are replaced by
those of the most distant of 50 random nodes, and \emph{clique}
anomalies, groups of 15 fully connected nodes. All methods share ten
labelled splits per benchmark, so comparisons are paired.

\paragraph{Graph-derived tables.}
\label{sec:tables}
Tabular detectors do not read the graph: it is first flattened, without
labels, into a table with one row per node. The tree-based comparisons use
GADBench's features, $\Phi=[X\,\|\,PX\,\|\,P^2X]$, where $X$ holds the
node attributes, $P$ averages a node's incoming neighbours (a zero row
when it has none) and $\|$ concatenates columns. TFM4GAD builds another
table, $[\,B\,\|\,\log d\,\|\,\log\pi\,\|\,U\,]$: $B$ is a neighbourhood
block, either a bank of Beta-wavelet filters of $X$ or the mean
aggregation $[X\,\|\,PX\,\|\cdots\|\,P^CX]$, with the type and
$C\in\{1,2\}$ fixed per dataset by the authors' configuration; $d$ is the degree,
$\pi$ the PageRank score and $U$ a 16-dimensional Laplacian positional
encoding (Appendix~\ref{app:ports}). Both tables are computed once on the
original graph and stay fixed. Island views recompute the added rows alone, on sampled
graph contexts and with the same feature construction; validation and
test nodes keep their canonical rows.

\paragraph{Resampling interface: allocation, transformation, mass.}
\label{sec:atm}
Let $V_L$ be the labelled training nodes, with labels $y_i\in\{0,1\}$
(1 for an anomaly), rows $\phi_i$ of the table $\Phi$ and weights $w_i$.
A resampling strategy is an operator $\mathcal R$ on the labelled
training set, which a detector $f$ then fits:
\[
(G,X)\longrightarrow\Phi\longrightarrow(\Phi_L,y_L,w)
\overset{\mathcal R}{\longrightarrow}
(\widetilde\Phi,\widetilde y,\widetilde w)\longrightarrow f .
\]
Fixing $\Phi$ and $f$ while $\mathcal R$ varies isolates the effect of
resampling. We describe $\mathcal R$ by three choices: \emph{allocation}
$\mathcal A$ selects the source rows, \emph{transformation}
$\mathcal T$ builds new rows from them, and the \emph{weight rule}
$\mathcal M$ sets the \emph{mass} of each class,
$\sum_{j:\widetilde y_j=c}\widetilde w_j$. They are experimental
controls, not independent components of every method
(Appendix~\ref{app:interface}).

\paragraph{Detectors.}
\label{sec:resultssetup}
Table~\ref{tab:design} lists the ten configurations.
\emph{Tree ensembles:} \textsc{XGBGraph} and \textsc{RFGraph} fit XGBoost
\citep{chen2016xgboost} and a random forest \citep{breiman2001rf} on
$\Phi$. We run each at GADBench's defaults and
after GADBench's random search (100 candidates, selected on validation
AUPRC, also over the number of hops and the aggregator;
Appendix~\ref{app:tuning}).
\emph{Tabular foundation model:} TFM4GAD gives TabPFN-v2 the 100 labelled
rows of its table as context and scores the test rows in context, with no
gradient step. To separate the model from the table we also
run TabPFN-v2 directly on $\Phi$.
\emph{GNNs:} BWGNN, GATv2 \citep{gatv2}, GHRN and SpaceGNN, untuned
(Appendix~\ref{app:ports}); our BWGNN, GHRN and SpaceGNN runs are within
0.6 AUPRC points of the published means at the same budget on seven
shared benchmarks (Appendix~\ref{app:anchors}). We call the first six
configurations \emph{tabular}. The configurations are not independent:
the tuned trees share their learner with the default trees, and GHRN is
BWGNN on a pruned graph.

\begin{table}[t]
\centering\small
\setlength{\tabcolsep}{4pt}
\caption{\textbf{Detector configurations and the strategies run in
each.} $\Phi$: neighbourhood-aggregated features (defined in the text).
CW: class weighting. Six samplers: ROS, ROSE, SMOTE, ADASYN, island views
and GODM. Unweighted training is run wherever class weighting is the
reference. ROSE restricted to the attributes $X$
(Section~\ref{sec:injected}) was also run in the default trees and the
four GNNs, and MGS in TabPFN on $\Phi$, the default trees and the four
GNNs.}
\label{tab:design}
\begin{tabularx}{\linewidth}{@{}l>{\raggedright\arraybackslash}p{3.4cm}>{\raggedright\arraybackslash}p{1.8cm}>{\raggedright\arraybackslash}X>{\raggedright\arraybackslash}p{2.9cm}@{}}\toprule
Family & Detector & Samplers act on & Configuration & Strategies (reference)\\\midrule
In-context & TFM4GAD \citep{tfm4gad} & rows of its graph table & authors', 6 of 17 datasets; default block elsewhere & ROS, ROSE (unresampled context)\\
\addlinespace[2pt]
& TabPFN-v2 on $\Phi$ & rows of $\Phi$ & default checkpoint; 500 random columns per member on wide tables & all six but ADASYN (unresampled context)\\
\addlinespace[5pt]
Trees on $\Phi$ & \textsc{XGBGraph}, \textsc{RFGraph}, default & rows of $\Phi$ & GADBench defaults & all six (CW)\\
\addlinespace[2pt]
& \textsc{XGBGraph}, \textsc{RFGraph}, tuned & rows of $\Phi$ & 100-candidate search & ROS, ROSE, SMOTE (CW)\\
\addlinespace[5pt]
GNN & BWGNN \citep{bwgnn} & hidden layer & GADBench default & all six (CW)\\
\addlinespace[2pt]
& GATv2 \citep{gatv2} & hidden layer & ours (Appendix~\ref{app:ports}) & all six (CW)\\
\addlinespace[2pt]
& GHRN \citep{ghrn} & hidden layer & GADBench default & ROS, ROSE, SMOTE, ADASYN (CW)\\
\addlinespace[2pt]
& SpaceGNN \citep{spacegnn} & hidden layer & authors', 7 of 17 datasets; code defaults elsewhere & ROS, ROSE, SMOTE, ADASYN (CW)\\
\bottomrule\end{tabularx}
\end{table}

\paragraph{Strategies.}
\label{sec:strategies}
All strategies change the labelled training set only, use no validation
or test label, and bring the 20 anomalies to parity with the 80 normals. \emph{Class weighting} reweights
the original rows (weights 4 and 1). The six \emph{samplers} add 60 rows:
\emph{ROS} duplicates anomalies; \emph{ROSE} \citep{rose} adds Gaussian
noise to each duplicate, with a per-feature standard deviation of 0.8 to
1.0 times that of the labelled anomalies (Appendix~\ref{app:interface};
other amplitudes in Appendix~\ref{app:bandwidth});
\emph{SMOTE} and \emph{ADASYN} interpolate between labelled anomalies;
\emph{island views} recompute an anomaly's features on a sampled subgraph
around it \citep{forestfiresampling}; and \emph{GODM} \citep{godm}
generates new attributed nodes. In these terms class weighting changes
$\mathcal M$ alone and the samplers reach the same mass, 80 per class,
with new rows: ROS, ROSE, SMOTE and island views allocate uniformly over
the anomalies and differ in $\mathcal T$; ADASYN also changes $\mathcal A$.
All resample the anomalies only:
applying ROSE's noise to the normals as well adds nothing
(Appendix~\ref{app:rose-sym}). A seventh, \emph{MGS}
\citep{sakho2026rebalancing}, was added last, in TabPFN on $\Phi$, the
default trees and the four GNNs: its default neighbourhood exceeds the 19 neighbours available, so its 60
rows come from one Gaussian fitted to the 20 labelled anomalies
(Appendix~\ref{app:mgs}). For trees and TabPFN the samplers act on
the rows of $\Phi$, and for TFM4GAD on the rows of its context table. For
GNNs they act on the hidden representation that enters the classifier,
recomputed at every epoch, with views and generated nodes encoded as
extra graphs (Appendix~\ref{app:ports}). At that level ROS is class
weighting with unequal integer weights (one plus the number of times an
anomaly is drawn), so its GNN column measures uneven weighting of the
anomalies, not new rows.

\paragraph{References and statistics.}
A sampler that only beats unweighted training may do no more than class
weighting, so the reference of trees and GNNs is class weighting, and we
also report unweighted training (Table~\ref{tab:cross-family}).
In-context models take no weights; their
reference is the unresampled context, which for TFM4GAD is the published
method where its code runs (Appendix~\ref{app:ports}). We report the test area under the
precision-recall curve (AUPRC) $\times100$.
Gains are computed within each split, averaged within each benchmark,
then equally over the benchmarks of a panel, with 95\% hierarchical
bootstrap intervals over benchmarks, then splits
(Appendix~\ref{sec:statistics}). We call a gain or a difference
\emph{resolved} when its interval excludes zero; when we write that
there is no gain, loss or difference, we mean that none is resolved, not
that the estimate is zero. The intervals treat the benchmarks as a
sample and are not adjusted for multiplicity: this paper reports several
hundred of them, so an isolated resolved gain is weak evidence, and
we rely on patterns that repeat across configurations. Only the
first default-tree comparison (class weighting, ROS, ROSE, SMOTE and
island views with both learners, GODM with XGBGraph) was fixed before its
test results were read; the two default-tree results the paper starts
from, the organic null and the injected ROSE gain, belong to it. Every
other experiment is exploratory, and most, including ADASYN, MGS, and
GODM with RFGraph, were added afterwards; no
model, configuration or checkpoint was selected on test data
(Appendix~\ref{sec:protocol}).

\section{Results}

\newcommand{\takeaway}[1]{\par\noindent\textbf{Takeaway.} #1}

\begin{table}[t]
\centering\small
\setlength{\tabcolsep}{3.2pt}
\caption{\textbf{Resampling across detector families (\Scarce{20}).} Ref.: test AUPRC ($\times100$) of the reference of each detector. Other columns: paired gain over that reference, averaged over ten splits, then datasets. Unw.: unweighted training. MGS: Multivariate Gaussian SMOTE. ROSE on $X$: the noise is applied to the node attributes only (trees: the block $X$ of $\Phi$; GNNs: synthetic input nodes, Section~\ref{sec:injected}). Bold: hierarchical-bootstrap 95\% interval excludes zero; ---: not run.}
\label{tab:cross-family}
\begin{tabular*}{\linewidth}{@{\extracolsep{\fill}}lrrrrrrr@{}}\toprule
Detector & Ref. & Unw. & ROS & ROSE & SMOTE & MGS & \shortstack{ROSE\\on $X$}\\
\midrule
\multicolumn{8}{c}{\textbf{Organic panel (8 datasets)}}\\
\addlinespace[2pt]
\multicolumn{8}{@{}l}{\emph{Tabular foundation model, labelled rows in context (gains over the unresampled context)}}\\
TFM4GAD & $46.3$ & --- & $-0.82$ & $-0.63$ & --- & --- & ---\\
TabPFN on $\Phi$ & $45.4$ & --- & $-1.29$ & $-0.25$ & $-1.39$ & $-0.78$ & ---\\
\addlinespace[2pt]
\multicolumn{8}{@{}l}{\emph{Tree ensembles on $\Phi$ (gains over class weighting)}}\\
\textsc{RFGraph}, tuned & $44.8$ & $-0.07$ & $-0.44$ & $-0.52$ & $-0.91$ & --- & ---\\
\textsc{XGBGraph}, tuned & $43.7$ & $+0.39$ & $-0.06$ & $+0.91$ & $+0.04$ & --- & ---\\
\textsc{RFGraph}, default & $41.5$ & $+1.24$ & $+0.23$ & $+0.99$ & $+0.52$ & $+0.79$ & $+0.06$\\
\textsc{XGBGraph}, default & $41.9$ & $-0.16$ & $-0.02$ & $+0.94$ & $+0.16$ & $+0.06$ & $-0.22$\\
\addlinespace[2pt]
\multicolumn{8}{@{}l}{\emph{Graph neural networks, samplers on the hidden representation (gains over class weighting)}}\\
SpaceGNN & $40.1$ & $+0.01$ & $\mathbf{-0.50}$ & $-0.39$ & $+0.24$ & $-0.08$ & $+0.01$\\
GHRN & $38.1$ & $-2.69$ & $-1.26$ & $-0.95$ & $+0.05$ & $-1.34$ & $-1.10$\\
BWGNN & $37.9$ & $-2.75$ & $-0.46$ & $-0.17$ & $+0.02$ & $-0.26$ & $-1.17$\\
GATv2 & $23.0$ & $-0.17$ & $-0.79$ & $-0.50$ & $-0.41$ & $+0.04$ & $-0.14$\\
\midrule
\multicolumn{8}{c}{\textbf{Injected panel (9 datasets)}}\\
\addlinespace[2pt]
\multicolumn{8}{@{}l}{\emph{Tabular foundation model, labelled rows in context (gains over the unresampled context)}}\\
TFM4GAD & $32.5$ & --- & $+1.75$ & $+0.25$ & --- & --- & ---\\
TabPFN on $\Phi$ & $29.1$ & --- & $+0.12$ & $\mathbf{+2.52}$ & $+1.18$ & $\mathbf{+2.24}$ & ---\\
\addlinespace[2pt]
\multicolumn{8}{@{}l}{\emph{Tree ensembles on $\Phi$ (gains over class weighting)}}\\
\textsc{RFGraph}, tuned & $29.4$ & $+0.54$ & $+0.64$ & $+1.64$ & $+0.73$ & --- & ---\\
\textsc{XGBGraph}, tuned & $25.1$ & $+0.45$ & $+1.05$ & $\mathbf{+2.57}$ & $\mathbf{+1.42}$ & --- & ---\\
\textsc{RFGraph}, default & $25.8$ & $+1.02$ & $+0.06$ & $\mathbf{+4.40}$ & $\mathbf{+1.76}$ & $\mathbf{+4.80}$ & $\mathbf{+4.14}$\\
\textsc{XGBGraph}, default & $20.7$ & $+0.80$ & $-0.49$ & $\mathbf{+5.22}$ & $\mathbf{+3.44}$ & $\mathbf{+4.68}$ & $\mathbf{+2.58}$\\
\addlinespace[2pt]
\multicolumn{8}{@{}l}{\emph{Graph neural networks, samplers on the hidden representation (gains over class weighting)}}\\
SpaceGNN & $16.7$ & $-0.03$ & $+0.75$ & $+0.67$ & $+0.41$ & $-0.15$ & $+0.83$\\
GHRN & $22.4$ & $\mathbf{-3.08}$ & $-0.21$ & $+0.02$ & $+0.20$ & $-0.08$ & $-0.14$\\
BWGNN & $22.8$ & $-2.20$ & $-0.02$ & $+0.00$ & $-0.12$ & $+0.18$ & $+0.06$\\
GATv2 & $18.3$ & $-0.44$ & $-0.46$ & $-0.25$ & $-0.02$ & $-0.67$ & $-0.65$\\
\bottomrule\end{tabular*}
\end{table}

\subsection{No statistically resolved average gain on eight organic-anomaly benchmarks}
\label{sec:rq1}

Table~\ref{tab:cross-family} gives, for each configuration, the gain of
unweighted training, ROS, ROSE and SMOTE over its reference; ADASYN,
island views and GODM are in Table~\ref{tab:app-cross-family-other}. On
the organic panel no gain in either table has an interval above zero. Two
have an interval below it (ROS in SpaceGNN, GODM in BWGNN), about what
this many unadjusted intervals produce by chance. The null is informative
for most configurations: ROSE's interval excludes gains above two points
in eight of ten (the default trees are the exceptions). Neither reference
dominates: unweighted training is above class weighting for default
RFGraph and below it for GHRN and BWGNN, none resolved.

\begin{table}[t]
\centering\small
\setlength{\tabcolsep}{4pt}
\caption{\textbf{ROSE on each organic benchmark} (gain over the reference of each detector, AUPRC points, mean of ten paired splits). Bold: 95\% bootstrap interval over the ten splits excludes zero; it resamples splits only, so it describes that benchmark and its ten splits (80 cells, unadjusted). Last column: datasets with a positive mean.}
\label{tab:organic-datasets}
\begin{adjustbox}{max width=\linewidth}
\begin{tabular}{lrrrrrrrrr}\toprule
Detector & Amazon & FDComp. & Questions & Reddit & T-Fin. & Tolokers & Weibo & YelpChi & $>0$\\\midrule
TFM4GAD & $-0.6$ & $\mathbf{-3.1}$ & $\mathbf{+1.5}$ & $\mathbf{+0.4}$ & $\mathbf{-1.4}$ & $-0.3$ & $\mathbf{-0.6}$ & $-1.0$ & 2/8\\
TabPFN on $\Phi$ & $\mathbf{-1.3}$ & $\mathbf{-1.9}$ & $+1.1$ & $\mathbf{-0.3}$ & $\mathbf{-2.8}$ & $\mathbf{-0.9}$ & $\mathbf{+0.8}$ & $\mathbf{+3.3}$ & 3/8\\
\addlinespace[2pt]
\textsc{RFGraph}, tuned & $-0.7$ & $+1.0$ & $+0.8$ & $+0.0$ & $\mathbf{-2.1}$ & $\mathbf{-2.2}$ & $-2.3$ & $+1.4$ & 4/8\\
\textsc{XGBGraph}, tuned & $+0.8$ & $+1.2$ & $\mathbf{+1.2}$ & $\mathbf{-0.3}$ & $+0.7$ & $-1.0$ & $\mathbf{+2.0}$ & $\mathbf{+2.7}$ & 6/8\\
\textsc{RFGraph}, default & $+1.0$ & $+0.9$ & $\mathbf{-1.4}$ & $\mathbf{-0.2}$ & $\mathbf{-2.8}$ & $\mathbf{-1.5}$ & $\mathbf{+7.0}$ & $\mathbf{+5.0}$ & 4/8\\
\textsc{XGBGraph}, default & $-0.6$ & $\mathbf{+1.4}$ & $+0.3$ & $-0.1$ & $+0.0$ & $\mathbf{-1.2}$ & $+3.6$ & $\mathbf{+4.1}$ & 5/8\\
\addlinespace[2pt]
SpaceGNN & $+0.0$ & $-0.1$ & $-0.2$ & $\mathbf{-0.3}$ & $\mathbf{-2.9}$ & $+0.0$ & $+0.5$ & $\mathbf{-0.2}$ & 3/8\\
GHRN & $-0.6$ & $-0.3$ & $\mathbf{+0.5}$ & $-0.2$ & $-5.1$ & $-0.4$ & $-0.7$ & $\mathbf{-0.9}$ & 1/8\\
BWGNN & $+1.1$ & $-0.1$ & $\mathbf{+0.6}$ & $-0.1$ & $-0.5$ & $-0.9$ & $-1.0$ & $-0.6$ & 2/8\\
GATv2 & $-2.4$ & $-0.2$ & $\mathbf{-0.6}$ & $\mathbf{+0.1}$ & $-0.1$ & $+0.4$ & $-1.4$ & $+0.2$ & 3/8\\
\bottomrule\end{tabular}\end{adjustbox}
\end{table}

\paragraph{The average hides opposite effects.}
Table~\ref{tab:organic-datasets} gives ROSE's gain on each organic
benchmark. No detector and no benchmark has a stable sign: T-Finance and
Tolokers lose with most configurations, while on YelpChi five of the six
tabular configurations gain and three of the four GNNs lose. The largest
gains, those of the default trees on Weibo, are fragile: RFGraph's is
measured against a weak reference (class weighting is 8.2 points below
unweighted training there, and ROSE stays below it), and XGBGraph's has
an interval over splits that includes zero.

\paragraph{Treating the sampler as a hyperparameter is safe and gains
little.}
A practitioner could select the sampler on validation data. Keeping, for
each split, the one of the reference and ROSE with the higher validation
AUPRC changes the organic score by $-0.3$ to $+1.4$ points over always
using the reference, resolved for default XGBGraph only
(Table~\ref{tab:app-selection}).

\paragraph{Splits matter as much as strategies.}
\label{sec:split-variation}
On each organic benchmark, a two-way decomposition of AUPRC over the ten
splits and the strategies attributes 53--79\% of the variation to the
splits and 3--16\% to the strategies
(Table~\ref{tab:app-variance-detectors}): the spread over splits is as
large as the range between the best and the worst strategy. Weibo is the
exception for default RFGraph, GHRN and BWGNN.

\takeaway{With twenty labelled organic anomalies, no strategy is reliably
better than its reference in any detector family; which nodes are
labelled matters more than how they are rebalanced.}

\begin{figure}[t]
\centering
\includegraphics[width=\linewidth]{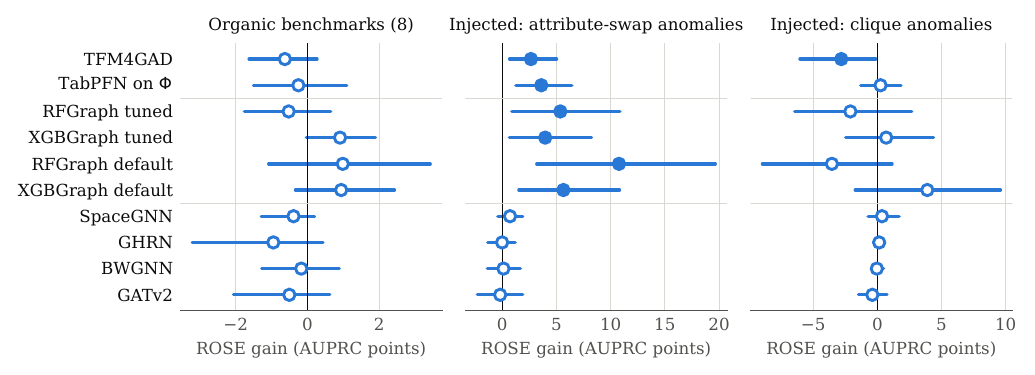}
\caption{\textbf{Where ROSE helps.} Gain of ROSE over the reference of
each configuration (AUPRC points, 95\% interval; class weighting, or the
unweighted method for the two in-context rows): on the organic panel,
and on the eight injected benchmarks with type labels, separately for
attribute-swap and clique anomalies. Filled markers: interval excludes
zero. Scales differ between panels.}
\label{fig:rose}
\end{figure}

\subsection{Injected anomalies: noise helps tabular detectors, consistently only on attribute swaps}
\label{sec:injected}

On the injected panel several samplers help the tabular detectors, ROSE
the most (Table~\ref{tab:cross-family}). Its gain is resolved in four of
the six tabular configurations, from $+2.5$ points (TabPFN on $\Phi$) to
$+5.2$ (default XGBGraph), and not in tuned RFGraph or TFM4GAD. SMOTE,
ADASYN and GODM gain less, each in some of these configurations
(Table~\ref{tab:app-cross-family-other}); MGS matches ROSE in the default
trees and TabPFN (differences unresolved). Duplication gives no gain, so
the gain comes from new feature values, not from added rows. In the four
GNNs no sampler has a gain, and three cells have an interval below zero
(ADASYN in GHRN and BWGNN, GODM in BWGNN). SpaceGNN comes closest: ROSE
is positive on all nine benchmarks, but unresolved.

\paragraph{By anomaly type.}
Eight injected benchmarks record the type of each anomaly. Scoring each
type against the normal test nodes (Figure~\ref{fig:rose},
Table~\ref{tab:app-types-detectors}) shows a gain on attribute swaps that
is resolved in all six tabular configurations, from $+2.6$ (TFM4GAD) to
$+10.8$ (default RFGraph), and in none of the four GNNs. It is carried by
five benchmarks; on Cora, Citeseer and Pubmed the swaps gain nothing. On
cliques no configuration gains and TFM4GAD loses. Default XGBGraph's
panel gain also contains large clique gains on those three citation
graphs ($+8.9$ to $+14.7$), most of which noise on the one-hop aggregate
alone reproduces (Table~\ref{tab:app-blocks}).

\paragraph{GNNs: hidden layer and input.}
Headroom does not explain the GNN null. Class-weighted BWGNN and GHRN
already score the swaps at 35 AUPRC, where the default trees score them
at 5 to 6, but SpaceGNN and GATv2 score them at 14 and 21 and show no
gain either. Nor does the place where the noise enters. We also add
ROSE's 60 synthetic anomalies, drawn on $X$ with the trees' sampler and
seeds, as input nodes that copy the incoming edges of their source
(Appendix~\ref{app:input-rose}). No change in a GNN's panel or swap score
is resolved (Table~\ref{tab:cross-family}, last column), and on cliques
GATv2 loses 0.9 points; for the trees the same restriction to $X$ keeps
most of RFGraph's gain and half of XGBGraph's. Retraining a
head with ROSE on frozen BWGNN or GraphSAGE embeddings gives one resolved
gain in twelve contrasts (Table~\ref{tab:app-frozen}).

\takeaway{On injected benchmarks the benefit of resampling is narrow: it
requires new feature values, a tabular detector and attribute-swap
anomalies; no placement of the noise helps a GNN.}

\subsection{The swap gain depends on the injection recipe}
\label{sec:leak}

The standard recipe copies, for each swap, the attributes of the most
distant of 50 random nodes. The most distant candidate tends to have a
large norm, the leak that \citet{huang2023unsupervised} documented for
unsupervised detection: here the norm alone scores the swaps
at 13 to 25 times their prevalence (Table~\ref{tab:app-labelfree}), 41
AUPRC on average; trees with ROSE reach 12 to 17. We
test whether ROSE's swap gain depends on this leak by changing the recipe
on graphs that we inject ourselves: three bag-of-words graphs
(Amazon-Photo, Amazon-Computers, Coauthor-CS) and six organic graphs with
real-valued features, used as hosts with their organic anomalies left
unlabelled (Appendix~\ref{app:recipe}).

\paragraph{A candidate ladder.}
\citet{huang2023unsupervised} attribute the leak to the number $k$ of
candidates. We rerun the standard recipe with $k=1$, 2, 5 and 15, keeping
the dense groups and the swap targets fixed (Table~\ref{tab:recipe}, top;
per graph in Table~\ref{tab:app-ladder}). With one candidate the donor is
a random node: the norm carries no signal, the swaps are close to chance
for the three detectors, and ROSE gains nothing. As $k$ grows, the norm
lift, the reference score and ROSE's gain rise together: for both trees
the gain is resolved from $k=15$ and reaches $+6.6$ and $+10.4$ points at
$k=50$. BWGNN reads the swaps better as $k$ grows and gains nothing from
ROSE. On Reddit and YelpChi the norm lift never exceeds $1.3\times$ and
the gain never exceeds $+1.5$; YelpChi's gain is unchanged under the
preserving recipe below, so a small part of it does not go through the
norm.

\begin{table}[t]
\centering\small
\setlength{\tabcolsep}{3.5pt}
\caption{\textbf{Attribute-swap anomalies under controlled injection recipes} (\Scarce{20}, AUPRC $\times100$, prevalence about 2.4\%). Ref.: class weighting (unweighted for TabPFN). $\Delta$: ROSE minus reference; bold: 95\% interval (graphs, injection realisations, splits) excludes zero. Norm lift: AUPRC of $\lVert x\rVert$ or $-\lVert x\rVert$ alone (the larger of the two in each split), divided by prevalence. Bag-of-words graphs: Amazon-Photo, Amazon-Computers, Coauthor-CS; hosts: FDCompCN, Questions, Reddit, Tolokers, Weibo, YelpChi. Ladder rows: two realisations per graph; other rows: three (two for the standard recipe on the bag-of-words graphs).}
\label{tab:recipe}
\begin{adjustbox}{max width=\linewidth}
\begin{tabular}{llrrrrrrrrr}\toprule
 & & & \multicolumn{2}{c}{\textsc{XGBGraph}} & \multicolumn{2}{c}{\textsc{RFGraph}} & \multicolumn{2}{c}{TabPFN on $\Phi$} & \multicolumn{2}{c}{BWGNN}\\
\cmidrule(lr){4-5}\cmidrule(lr){6-7}\cmidrule(lr){8-9}\cmidrule(lr){10-11}
Graphs & Recipe & Norm lift & Ref. & $\Delta$ & Ref. & $\Delta$ & Ref. & $\Delta$ & Ref. & $\Delta$\\\midrule
All nine & Standard, 1 candidate & $1.1\times$ & $2.8$ & $-0.1$ & $2.9$ & $-0.1$ & --- & --- & $3.0$ & $+0.1$\\
 & Standard, 2 candidates & $1.3\times$ & $3.1$ & $+0.1$ & $3.2$ & $+0.0$ & --- & --- & $3.8$ & $+0.1$\\
 & Standard, 5 candidates & $2.3\times$ & $3.9$ & $+0.6$ & $4.4$ & $+0.9$ & --- & --- & $7.5$ & $+0.4$\\
 & Standard, 15 candidates & $5.3\times$ & $6.1$ & $\mathbf{+2.5}$ & $6.2$ & $\mathbf{+4.8}$ & --- & --- & $16.4$ & $+0.3$\\
 & Standard, 50 candidates & $13.0\times$ & $11.4$ & $\mathbf{+6.6}$ & $12.6$ & $\mathbf{+10.4}$ & --- & --- & $30.8$ & $+1.3$\\
\addlinespace[3pt]
Bag-of-words (3) & Standard (50 candidates) & $17.3\times$ & $13.4$ & $\mathbf{+11.6}$ & $11.5$ & $\mathbf{+19.8}$ & $23.5$ & $\mathbf{+5.8}$ & $55.8$ & $+2.4$\\
 & Norm- and degree-preserving & $1.1\times$ & $2.6$ & $+0.1$ & $2.6$ & $-0.0$ & $2.6$ & $+0.1$ & $3.1$ & $-0.1$\\
 & \emph{Standard $-$ preserving} &  & $\mathbf{+10.8}$ & $\mathbf{+11.5}$ & $\mathbf{+9.0}$ & $\mathbf{+19.8}$ & $\mathbf{+21.0}$ & $\mathbf{+5.7}$ & $\mathbf{+52.7}$ & $+2.4$\\
\addlinespace[3pt]
Hosts (6) & Standard (50 candidates) & $10.9\times$ & $10.8$ & $\mathbf{+3.6}$ & $13.0$ & $\mathbf{+5.9}$ & $17.6$ & $+2.4$ & $18.3$ & $+0.3$\\
 & Norm- and degree-preserving & $1.1\times$ & $4.4$ & $+0.6$ & $4.4$ & $+0.4$ & $8.6$ & $-2.7$ & $4.4$ & $+0.0$\\
 & \emph{Standard $-$ preserving} &  & $\mathbf{+6.5}$ & $\mathbf{+3.1}$ & $\mathbf{+8.6}$ & $\mathbf{+5.6}$ & $\mathbf{+9.0}$ & $\mathbf{+5.1}$ & $\mathbf{+13.9}$ & $+0.3$\\
\bottomrule\end{tabular}\end{adjustbox}
\end{table}

\paragraph{A norm-preserving swap.}
A second variant keeps the 50 candidates but draws them among the nodes
of closest norm, and builds the dense groups by degree-preserving edge
swaps (Table~\ref{tab:recipe}). Under the standard recipe ROSE's swap
gain is resolved in five of six cases (XGBGraph, RFGraph and TabPFN, on
the bag-of-words graphs and on the hosts) and reaches $+19.8$ points.
Under the preserving recipe the norm carries no signal and no gain is
resolved; the difference between recipes is resolved in all six cases.
The test has a limit: removing the norm signal also makes the swaps hard
to detect. On the bag-of-words graphs every detector falls to chance,
BWGNN included. The clearest case above the floor is the Reddit host, where
TabPFN scores the swaps at about 20 AUPRC under both recipes although the
norm carries almost nothing: there ROSE lowers the score, by 8 points
under the standard recipe and 15 under the preserving one
(Table~\ref{tab:app-recipe-graphs}). Without the leak ROSE is not neutral
on the bag-of-words graphs: its overall effect is negative for both trees
($-1.1$, $-3.3$; resolved), via the dense groups.

\paragraph{What the tests support.}
On the nine graphs where we control the recipe, ROSE's swap gain tracks
the norm leak: it grows with it along the ladder, and without it no gain
is resolved and no estimate exceeds two points on any graph.
Detectability follows the same path, so neither test separates the two.
Where swaps stay readable without the leak (the Reddit host; FDCompCN and
Tolokers, at two to four times prevalence), ROSE does not recover the
gain. A leak is not sufficient either: Cora, Citeseer and Pubmed leak 14
to 25 times prevalence and gain nothing on swaps. Three controls on the
default trees narrow what the noise must do: extending it to the labelled
normals adds nothing (Appendix~\ref{app:rose-sym}); rows confined to the
span of the 20 labelled anomalies (MGS) match ROSE on swaps
(Table~\ref{tab:app-mgs}); and one perturbed row repeated for all copies
of a source keeps most of the gain (Table~\ref{tab:app-static-controls}).
Beyond this we do not know which property of the noise matters.

\paragraph{Two label-free columns.}
If the gain recovers signal that the detector fails to read, giving the
signal directly should compete with it. Appending the in-degree and the
feature norm to $\Phi$ raises class-weighted XGBGraph by $+8.1$ points on
the injected panel, more than ROSE does, and halves ROSE's gain, which is
no longer resolved (its swap gain largely remains in point estimate;
Table~\ref{tab:app-leak-columns}). ROSE also perturbs the two columns and
RFGraph barely uses them, so this is consistent with ROSE partly standing
in for information XGBGraph could be given directly, without
establishing it.

\paragraph{The panel contrast is not reproduced at fixed graph.}
The organic and injected panels use different graphs. We therefore
inject the standard recipe into five organic graphs and compare ROSE's
gain on the injected and the organic task of the same graph
(Table~\ref{tab:app-samegraph}). For
XGBGraph the difference between the two tasks is unresolved ($+0.6$, with
an interval that still contains the difference between panels, $+4.3$);
for RFGraph it is reversed ($-6.0$, resolved), through losses on cliques
that disappear on one realisation with cliques sized to the host's
degree. The five graphs are not the organic panel either: XGBGraph's
organic gain is resolved on them, because Weibo and YelpChi are among
them and Amazon, T-Finance and Tolokers are not, so the organic null is a
statement about the eight-benchmark panel. We therefore do not attribute
the gain to injection as such; what replicates on these graphs is the
gain on attribute swaps ($+3.0$ and $+5.2$, both resolved).

\takeaway{On the standard injected benchmarks, ROSE's gain rises and
falls with a label-free artefact of the recipe that also decides whether
the swaps are detectable. A gain there may owe as much to the recipe as
to the sampler; it should not be read as a property of injected
anomalies.}

\subsection{Detector choice and tuning matter as much as the sampler}
\label{sec:tuning}

The reference columns of Table~\ref{tab:cross-family} rank the
configurations. With no model tuned, RFGraph is above each of six GNN detectors (BWGNN,
GHRN, SpaceGNN, GraphSAGE, PC-GNN and ConsisGAD) on average, with a
resolved margin against three of them over the 17 benchmarks and against
one on the organic panel; TFM4GAD has one against each of the six on
both (Table~\ref{tab:app-margins}). The ranking is not uniform: the best
GNN is above tuned RFGraph on five of the nine injected graphs, by up to
21 points, although on that panel the label-free norm score alone (25.6)
exceeds every GNN mean (at most 22.8; Table~\ref{tab:app-leaderboard-inj}).

Over the 17 benchmarks the random search adds about three points to each
tree detector, as much as ROSE adds to the default detectors. Once they
are tuned, ROSE's gain falls from $+3.2$ to $+1.8$ points for XGBGraph
(still resolved) and from $+2.8$ to $+0.6$ for RFGraph
(Table~\ref{tab:app-tuning-effect}). On the organic panel tuning adds
$+1.8$ and $+3.3$ points, the first resolved, against about one for
ROSE. Three caveats apply.
The search also selects the aggregation, so it changes the representation
as well as the learner. It uses the 100 validation labels, which the
default tree detectors never see (the GNNs use them to select the
epoch). And only the trees were tuned: SpaceGNN's
own tuned results on the shared benchmarks are above our run of it
(Appendix~\ref{app:anchors}).

\takeaway{A validation search is worth as much as the best sampler and
removes about half of its gain or more: tune the detector before crediting a
sampler.}

\subsection{Label budget: label count versus class ratio}
\label{sec:budget}

\Scarce{20} fixes the number of labels and their 1:4 ratio together. We
therefore cross 5, 10 and 20 labelled anomalies with 20, 40 and 80
labelled normals on three organic and three injected benchmarks, with
nested training labels and the same test nodes in the nine cells (default
trees; Table~\ref{tab:app-grid}). \emph{More labels at a fixed ratio:}
from 5 anomalies and 20 normals to 20 and 80, four times the labels at
the same 1:4 ratio, ROSE's gain over class weighting does not change for
either tree (the contrast we fixed before running the grid).
\emph{More anomalies at a fixed number of normals:} on the injected
benchmarks the gain falls, in three of four comparisons; it is resolved
in every cell with five anomalies and in none with twenty. On the organic
benchmarks no cell and no comparison is resolved, although with five
anomalies XGBGraph's estimates reach $+4.0$ to $+5.4$ points with wide
intervals. \emph{Full supervision:} on GADBench's fully supervised masks
(other test nodes, means only), ROSE's organic gain over class weighting
is $+0.6$ and $-0.0$ for the two trees. Its injected gain is $+2.8$ and
$+4.1$ over class weighting, but only $+0.9$ and $-0.4$ over unweighted
training, which is the stronger reference there
(Tables~\ref{tab:app-fullsup-xgb} and~\ref{tab:app-fullsup-rf}).

\takeaway{No label budget we tried gives a resolved organic gain. The
injected gain is largest where labelled anomalies are fewest; with six
benchmarks we cannot attribute this to their number or to the class
ratio (Appendix~\ref{app:budget}).}

\subsection{Matched \texorpdfstring{$\mathcal A/\mathcal T/\mathcal M$}{A/T/M} controls: graph views against copies}
\label{sec:rq2}
\label{sec:adaptive-resampling}

Island views did not beat class weighting above, and GODM beat its
reference only on the injected panel (default XGBGraph and TabPFN on
$\Phi$). But a sampler compared with class weighting may differ from it
in $\mathcal A$, $\mathcal T$ and $\mathcal M$ at once
(Section~\ref{sec:atm}). We fix allocation and mass and vary the
transformation alone. A warm-up XGBoost model scores the labelled nodes of both classes by
residual; 50 sources are drawn with a probability that increases with the
residual and oversampled as exact copies, as interpolations (SMOTE on
$\Phi$) or as island views (a bounded variant of the sampler), with class
mass either redistributed over old and new rows or added
(Appendix~\ref{app:train}). Sources, residuals and splits are identical
across arms, and each arm's rows are then given unchanged to RFGraph.

\begin{table}[t]
\centering\small
\setlength{\tabcolsep}{3pt}
\caption{\textbf{Matched controls} (paired gain in AUPRC points, 95\%
interval below; bold: interval excludes zero). Each contrast changes one
component and holds the selected sources, the splits and the learner
fixed. Allocation always uses the XGBoost residuals. \textsc{XGBGraph} here
uses 200 trees and learning rate 0.05 (Appendix~\ref{app:xgb-configurations}).
RFGraph: one forest on the original and generated rows (pooled);
Table~\ref{tab:app-rf-adaptive} gives a second design in which the forest
follows the booster's block schedule (blockwise).}
\label{tab:matched}
\begin{tabular}{@{}llcccc@{}}\toprule
& & \multicolumn{2}{c}{\textsc{XGBGraph}} & \multicolumn{2}{c}{\textsc{RFGraph}}\\
\cmidrule(lr){3-4}\cmidrule(lr){5-6}
Component changed & Contrast & Organic (8) & Injected (9) & Organic (8) & Injected (9)\\\midrule
Transformation $\mathcal T$ & Views $-$ copies & $+0.28$ & $-0.32$ & $+0.87$ & $+0.15$\\
 & & {\scriptsize$[-0.18,+0.94]$} & {\scriptsize$[-1.40,+0.59]$} & {\scriptsize$[-0.34,+2.41]$} & {\scriptsize$[-0.69,+0.90]$}\\[2pt]
Transformation $\mathcal T$ & SMOTE $-$ copies & $+0.23$ & $\mathbf{+0.96}$ & $-0.21$ & $\mathbf{+1.68}$\\
 & & {\scriptsize$[-0.12,+0.63]$} & {\scriptsize$[+0.45,+1.49]$} & {\scriptsize$[-0.86,+0.35]$} & {\scriptsize$[+0.89,+2.49]$}\\[2pt]
Mass $\mathcal M$ & Added $-$ fixed (views) & $+0.02$ & $-0.09$ & $\mathbf{-1.26}$ & $\mathbf{-2.26}$\\
 & & {\scriptsize$[-0.29,+0.40]$} & {\scriptsize$[-0.68,+0.50]$} & {\scriptsize$[-2.92,-0.13]$} & {\scriptsize$[-4.73,-0.32]$}\\[2pt]
Allocation $\mathcal A$ & Residual $-$ class weights & $\mathbf{+0.20}$ & $\mathbf{+0.33}$ & $+0.23$ & $+0.04$\\
 & & {\scriptsize$[+0.02,+0.41]$} & {\scriptsize$[+0.05,+0.59]$} & {\scriptsize$[-0.12,+0.71]$} & {\scriptsize$[-0.42,+0.57]$}\\
\bottomrule\end{tabular}
\end{table}

Replacing copies by island views of the same nodes has no resolved effect
in either panel, with either learner (Table~\ref{tab:matched}). The
comparison can detect a transformation effect on the injected panel,
where interpolation beats copies with both learners; on the organic panel
we have no positive control. Adding class mass does not help; it hurts
RFGraph in the one design that adds more mass than the XGBoost control
(Table~\ref{tab:app-rf-adaptive}).
Allocating by residual gives XGBGraph a small resolved gain and RFGraph
none. Two results point the other way: every RFGraph estimate of views
minus copies is positive, and resolved in the blockwise design over all
17 benchmarks ($+0.44$); and with RFGraph views beat class weighting on
the organic panel ($+1.40$).

\takeaway{At matched allocation and class mass, island views show no
resolved benefit over plain copies within either panel, which is not
evidence that they have none. Compare a new graph sampler with copies of
the same nodes before attributing its gain to the generated structure.}

\section{Limitations}

\emph{Detectors.} Samplers were not inserted into ConsisGAD or PC-GNN,
which carry their own augmentation or neighbour sampling, and PC-GNN runs
as GADBench's full-batch simplification. SpaceGNN and TFM4GAD 
ran outside their published configuration on
the nine injected benchmarks and respectively on one and three 
organic ones, and SpaceGNN's reference uses a class-weighted loss that
the published model does not.
Among the GNNs, island views and GODM were run in BWGNN and GATv2 only, and the matched
controls rely on boosting residuals and cover trees only.
\emph{Tuning.} Only the tree detectors were tuned on each dataset; the
GNNs and the in-context models were not (Appendix~\ref{app:tuning}).
\emph{In-context models.} Oversampling TFM4GAD's context changes how
TabPFN types some columns on six benchmarks; TabPFN on $\Phi$ types all
columns as numeric, but reads random subsets of 500 columns on the eleven
benchmarks with wider tables.
\emph{Injection.} The recipe tests use nine graphs and two families of
variants, with three detector families for the preserving recipe and two
(trees and BWGNN) for the candidate ladder, on sparse graphs a
degree-preserving group is far from a clique.
\emph{Statistics.} As an empirical study with eight organic datasets 
we have limited statistical precision. Intervals containing zero 
do not exclude meaningful gains.

\section{Recommendations and Conclusion}

Across ten detector configurations from three families, none of the six
samplers had a statistically resolved average gain on the eight
organic-anomaly benchmarks with twenty labelled anomalies. The clearest gain, Gaussian
noise on injected benchmarks, is confined to tabular detectors and
attribute swaps; it rises and falls with a label-free artefact of the
injection recipe, which our tests do not separate from the detectability
of the swaps. Known in unsupervised detection
\citep{huang2023unsupervised}, this artefact also decides where samplers
appear to help.
Tuning the trees brought as much as ROSE.

\noindent\textbf{Practical guidance.}
\emph{Tune baselines:} tune the detector before crediting a sampler, and
compare class weighting with unweighted training, since neither
dominates. \emph{Repeat labelled splits:} keep them paired and report
their variation; with eight organic benchmarks and ten paired splits, the
interval of ROSE's gain has a half-width of 0.7 to 2.3 points, so a claim
of one to two AUPRC points needs more splits or benchmarks.
\emph{Distinguish organic from injected anomalies:} report the two
separately, and injected anomalies by type. \emph{Check degree and norm
baselines:} on injected benchmarks, report the label-free degree and
feature-norm scores \citep{huang2023unsupervised}; a gain on a type that
they already detect is weak evidence for the sampler. With about twenty
labelled organic anomalies, treat resampling as a hyperparameter at most:
selecting ROSE or the reference on validation data gained at most 1.4
points.

\noindent\textbf{For evaluating a new sampler,} we add three checks.
(i) Test the mechanism: compare generated rows with copies of the same
sources at the same class mass. (ii) Vary the label count and the class
ratio separately. (iii) Test one detector
from another family: perturbing rows of $\Phi$ gave no gain
inside any GNN, at the hidden layer or at the input.

\subsubsection*{Acknowledgments}
Generative AI tools were used to assist with drafting and editing the
manuscript, to improve its clarity and readability, and with writing and
reviewing the code of the experiments. The authors take full
responsibility for the final content of the paper and of the code,
including all text, results and claims produced with AI assistance.

\clearpage
\bibliographystyle{iclr/format_iclr/iclr2027_conference}
\IfFileExists{reference.bib}
  {\bibliography{reference}}
  {\bibliography{paper/reference}}

\appendix

\section{Resampling Interface and Matched Controls}
\label{app:interface}
ROSE is the smoothed bootstrap of \texttt{imbalanced-learn}
\citep{imblearn} with shrinkage one: a synthetic row is a labelled
anomaly plus Gaussian noise of standard deviation
$h_j=(4/((d+2)n))^{1/(d+4)}\,\sigma_j$ in feature $j$, with $\sigma_j$
the standard deviation of that feature over the 20 labelled anomalies,
$n=100$ labelled training rows and $d$ features. The factor is 0.82 for
the 30 columns of T-Finance and Tolokers, 0.91 to 0.96 on Amazon,
YelpChi, FDCompCN and Reddit, and 0.99 or more on Questions, Weibo and
the nine injected benchmarks, so $h_j\approx\sigma_j$ on the wide tables;
features that are constant among the labelled anomalies receive no noise
(other windows: Appendix~\ref{app:bandwidth}).

\paragraph{Island views.}
For a labelled anomaly $v$, an \emph{island view} is the row
$\phi(v;G_v)$ recomputed, with the feature construction of the canonical
row $\phi_v=\phi(v;G)$, on a connected induced subgraph $G_v\ni v$
sampled by a seeded forest-fire burn \citep{forestfiresampling}. Each
view is an additional training row with the label of its source;
validation and test nodes keep their canonical rows. Static views use the
unbounded sampler and the original edge directions. The residual-guided
views of Section~\ref{sec:rq2} use the bounded Bernoulli variant
\textsc{IslandBurn} (Appendix~\ref{app:ff}), sampled and encoded on a
loop-free bidirected copy of $G$.

\paragraph{Matched controls.}
The controls of Section~\ref{sec:rq2} separate three choices:
\emph{allocation} $\mathcal A$ selects the source rows,
\emph{transformation} $\mathcal T$ builds the training rows, and the
\emph{weight rule} $\mathcal M$ assigns their weights.
\emph{Allocation:} an XGBoost warm-up predictor gives probabilities
$p_i^{(\mathrm w)}$ and residual hardness $|y_i-p_i^{(\mathrm w)}|$,
frozen after warm-up, so allocation is the same for every transformation
and receives no feedback from generated rows.
\emph{Transformation:} the selected sources are added as unchanged
copies, as island views, or as interpolations
$\widetilde\phi=\phi_i+\lambda(\phi_j-\phi_i)$, where $j$ is drawn
uniformly from up to five nearest labelled training neighbours of the same
class in $\Phi$ and $\lambda\sim\mathcal U(0,1)$; the row keeps $y_i$, and
a source without a same-class neighbour is duplicated.
\emph{Weight rule:} with $V_L$ the labelled training nodes, $w_i$ their
weights before resampling and $\widetilde w_j$ the weights of all
resulting rows, the primary rule redistributes the mass of each class
over its original and generated rows:
\[
\sum_{j:\widetilde y_j=c}\widetilde w_j
=\sum_{i\in V_L:y_i=c}w_i,\qquad c\in\{0,1\}.
\]
The added-mass control keeps the original weights and gives each generated
row its source's full weight. The adaptive comparisons share sources,
resampling budgets and boosting schedules (Appendix~\ref{app:train}).

\section{Cross-Family Experiments}
\label{app:cross-family-setup}

All experiments of this section are exploratory
(Appendix~\ref{sec:protocol}).

\subsection{TabPFN on \texorpdfstring{$\Phi$}{Phi}}
TFM4GAD changes the model and the table at once. To isolate the model,
TabPFN-v2 (checkpoint and 8-member ensemble of our TFM4GAD port) takes
the 100 labelled rows of $\Phi$ as context and scores the test rows.
Samplers act on the context rows with the trees' seeds, so a context
holds at most 160 rows; TabPFN takes no observation weights, so the
reference is the unresampled context. On tables wider than 500 columns
(Weibo, Questions and the nine injected datasets) each ensemble member
reads a random subset of 500 columns.

\subsection{Samplers inside GHRN and SpaceGNN}
GHRN is BWGNN trained on the graph pruned by GHRN's preprocessing, with
the arms of BWGNN (Appendix~\ref{app:ports}). In SpaceGNN each arm is
applied to every mini-batch at the input of the classifier of each of its
three towers, with one seed for the three; SMOTE and ADASYN search
neighbours in each tower's own representation. Both detectors ran ROS,
ROSE, SMOTE and ADASYN, but not island views or GODM, whose added graphs
GHRN's edge pruning and SpaceGNN's batching would handle differently from
the published models.

\subsection{ROSE at the input of a GNN}
\label{app:input-rose}
Hidden-layer ROSE perturbs a learned representation, whereas the trees'
ROSE perturbs fixed features. To put the noise on fixed attributes, we
draw ROSE's 60 synthetic anomalies on the attributes $X$ of the 100
labelled nodes, with the sampler and seeds of the trees, and add them to
the graph as labelled nodes. Each receives the incoming edges of its
source anomaly and sends no message, so the representation of every
original node is unchanged. The new nodes enter the loss as anomalies,
which brings the classes to parity, and are never scored. In GHRN they
are added after the edge pruning. In SpaceGNN they join the labelled
training set: an epoch has three mini-batches of 50 instead of two, and
parity holds over the training set, not within each batch. For the
default trees, ``ROSE on $X$'' in Table~\ref{tab:cross-family} restricts
the noise to the block $X$ of $\Phi$ (Table~\ref{tab:app-blocks}).

\subsection{ROSE on frozen GNN embeddings}
A BWGNN or GraphSAGE encoder is trained on the split and frozen, and a
head (XGBoost, random forest or MLP) is retrained on its embeddings with
class weighting or ROSE (Table~\ref{tab:app-frozen}). One of the twelve
contrasts is resolved: the XGBoost head on BWGNN embeddings on the
injected panel ($+1.5$ $[+0.3,+2.7]$).

\subsection{A norm- and degree-preserving injection recipe}
\label{app:recipe}
The standard recipe \citep{dominant,cola} makes a contextual anomaly by
copying the attributes of the most distant of 50 uniformly drawn nodes, and
a structural anomaly by wiring 15 nodes into a clique. Both leave a
label-free trace \citep{huang2023unsupervised}: a larger degree, and a
shifted feature norm (larger on six of our nine graphs, smaller on
Tolokers, unchanged on YelpChi and Reddit; the norm lift of
Table~\ref{tab:recipe} takes either sign). Following suggestions of
\citet{huang2023unsupervised}, our variant keeps the anomaly counts, the
half-and-half split and the group size, and changes two steps.
\emph{Attribute swaps:} the 50 candidates are the nodes whose $\ell_2$
norm is closest to the target's, and the most distant of them is copied,
so the norm barely moves (median ratio after to before of 1.00 on eight
graphs and 1.01 on Weibo). \emph{Dense groups:} the 15 members are
connected by double edge swaps, $(a,x),(b,y)\to(a,b),(x,y)$ with $a,b$ in
the group and $x,y$ outside, so every node keeps its degree exactly. A
group becomes as dense as the degrees of its members allow: on average,
98\% of the within-group pairs are linked on YelpChi and Weibo, about
70\% on Tolokers and the two Amazon graphs, 42\% on Coauthor-CS, and 6\%,
15\% and 35\% on FDCompCN, Questions and Reddit.

Both recipes are applied, in three realisations each (two for the
standard recipe on the bag-of-words graphs, where the third is the
benchmark file itself), to the three support graphs that we inject
ourselves and to six organic graphs. On the organic graphs the edges are
symmetrised first, anomalies are injected on 5\% of the nodes that are
not organic anomalies, and the organic anomalies stay as context but are
never injected, drawn as candidates, labelled or scored. Default
\textsc{XGBGraph} and \textsc{RFGraph}, TabPFN on $\Phi$ and BWGNN run on
the ten \Scarce{20} splits of each realisation
(Tables~\ref{tab:app-recipe} and~\ref{tab:app-recipe-graphs}).

\subsection{A candidate ladder for the standard recipe}
\label{app:ladder}
\citet{huang2023unsupervised} trace the norm leak to the number $k$ of
candidates: the most distant of $k$ uniformly drawn nodes tends to have a
larger norm when $k$ is larger. We rerun the standard recipe with
$k\in\{1,2,5,15\}$ next to the standard $k=50$, on the same nine graphs,
in two realisations each. Within a realisation the dense groups and the
swap targets are the same for every $k$; only the donor of each swap
changes (with $k=1$, a uniformly drawn node). Default \textsc{XGBGraph},
\textsc{RFGraph} and BWGNN run with class weighting and with ROSE on the
ten \Scarce{20} splits of each realisation (Table~\ref{tab:app-ladder}).

\subsection{Why only the anomalies are resampled}
\label{app:rose-sym}
Every sampler in this paper adds rows to the anomaly class only. To check
that nothing is lost by leaving the 80 labelled normals untouched, we
applied ROSE's kernel to them, scaled by the normals' own standard
deviations, in three ways: ROSE plus 80 noisy bootstrap draws of the
normals; ROSE plus 240 draws, so that three quarters of the rows are
noisy in both classes; and 240 draws with the anomalies only reweighted.
The weights of the normal rows are divided so that both classes keep the
same total weight in every arm (default trees, 17 benchmarks;
Table~\ref{tab:app-rose-sym}).

Noise on the normals alone gains in no cell and is resolved below class
weighting for \textsc{RFGraph} on both panels. Added to ROSE, it is never
distinguishable from ROSE (with 240 draws on the injected panel, $-1.80$
$[-4.56,+0.56]$ for \textsc{XGBGraph} and $-0.06$ $[-4.47,+4.22]$ for
\textsc{RFGraph}), and its point estimate is lower in seven of the eight
cells. The gain therefore comes with the anomaly rows, and resampling the
normals costs up to four times as many rows for no measured benefit.

\begin{table}[!htbp]
\centering\small
\setlength{\tabcolsep}{4pt}
\caption{\textbf{ROSE's kernel on the normal class} (\Scarce{20}, default trees, AUPRC points over class weighting, 95\% interval). Bold: interval excludes zero.}
\label{tab:app-rose-sym}
\begin{adjustbox}{max width=\linewidth}
\begin{tabular}{lllll}\toprule
 & \multicolumn{2}{c}{\textsc{XGBGraph}} & \multicolumn{2}{c}{\textsc{RFGraph}}\\\cmidrule(lr){2-3}\cmidrule(lr){4-5}
Noisy rows & Injected (9) & Organic (8) & Injected (9) & Organic (8)\\\midrule
Anomalies only (ROSE) & $\mathbf{+5.22}$ $[+2.69,+7.44]$ & $+0.94$ $[-0.33,+2.43]$ & $\mathbf{+4.40}$ $[+0.48,+8.53]$ & $+0.99$ $[-1.08,+3.42]$\\
ROSE $+$ 80 normal draws & $\mathbf{+3.50}$ $[+0.09,+6.56]$ & $+0.79$ $[-0.61,+2.50]$ & $+5.06$ $[-0.33,+10.88]$ & $+0.37$ $[-2.25,+3.11]$\\
ROSE $+$ 240 normal draws & $\mathbf{+3.42}$ $[+0.37,+6.23]$ & $+0.43$ $[-1.13,+2.18]$ & $+4.33$ $[-1.66,+10.31]$ & $-0.09$ $[-2.85,+2.56]$\\
Normals only (240 draws) & $-1.08$ $[-3.70,+1.60]$ & $+0.03$ $[-0.97,+1.00]$ & $\mathbf{-3.79}$ $[-7.43,-0.19]$ & $\mathbf{-3.99}$ $[-8.22,-0.53]$\\
\bottomrule\end{tabular}\end{adjustbox}
\end{table}

\subsection{Static ROSE controls: weights, copies, fixed and independent noise}
\label{app:static-controls}
Table~\ref{tab:app-static-controls} decomposes ROSE in the default trees
with the interface of Section~\ref{sec:atm}. All arms use the sources
that ROSE draws and, except the last, keep a mass of 80 per class.
\emph{Matched source weights} adds no row and gives each anomaly the
weight of its copies ($\mathcal A$ alone); \emph{copies} adds them as
rows (ROS); \emph{fixed perturbation} perturbs each source once and
repeats that row for all its copies; \emph{independent noise} is ROSE;
\emph{added mass} gives ROSE's rows the class weight as well, which
raises the anomaly mass from 80 to 320 on identical rows ($\mathcal M$
alone). Weights and copies are equivalent for \textsc{XGBGraph} (within
$10^{-5}$ points) and not distinguishable for \textsc{RFGraph}. On the
injected panel a fixed perturbation gains $+4.94$ $[+2.52,+7.12]$ over
copies for \textsc{XGBGraph} and $+3.52$ $[-0.35,+7.09]$ for
\textsc{RFGraph}; independent draws add $+0.77$ $[-0.04,+1.53]$ and
$+0.81$ $[-0.11,+1.79]$ to it. On the organic panel none of these
contrasts is resolved. Most of the injected gain therefore comes from
moving each anomaly once, not from the diversity of the draws or from
added mass.

\begin{table}[!htbp]
\centering\small
\setlength{\tabcolsep}{3.5pt}
\caption{\textbf{Static ROSE controls} (\Scarce{20}, default trees). Test
AUPRC and gain $\Delta$ over class weighting, not over the preceding row
(both $\times100$); parentheses count the datasets with a positive gain.
Intervals of the contrasts between rows are in the text.}
\label{tab:app-static-controls}
\begin{adjustbox}{max width=\linewidth}
\begin{tabular}{llrrrrrrrr}\toprule
Component & Variant & \multicolumn{4}{c}{\textsc{XGBGraph}} & \multicolumn{4}{c}{\textsc{RFGraph}}\\
\cmidrule(lr){3-6}\cmidrule(lr){7-10}
& & \multicolumn{2}{c}{Organic (8)} & \multicolumn{2}{c}{Injected (9)}
& \multicolumn{2}{c}{Organic (8)} & \multicolumn{2}{c}{Injected (9)}\\
& & AUPRC & $\Delta$ (wins) & AUPRC & $\Delta$ (wins)
& AUPRC & $\Delta$ (wins) & AUPRC & $\Delta$ (wins)\\\midrule
Reference & Unweighted & $41.78$ & $-0.16$ (3) & $21.54$ & $+0.80$ (6) & $42.76$ & $+1.24$ (4) & $26.86$ & $+1.02$ (6)\\
& Class weighting & $41.94$ & --- & $20.74$ & --- & $41.52$ & --- & $25.84$ & ---\\
\addlinespace[3pt]
Allocation $\mathcal A$ & Matched source weights & $41.92$ & $-0.02$ (5) & $20.25$ & $-0.49$ (3) & $41.57$ & $+0.06$ (3) & $25.97$ & $+0.13$ (5)\\
& Copies (ROS) & $41.92$ & $-0.02$ (5) & $20.25$ & $-0.49$ (3) & $41.74$ & $+0.23$ (5) & $25.90$ & $+0.06$ (5)\\
\addlinespace[3pt]
Transformation $\mathcal T$ & Fixed perturbation & $42.30$ & $+0.36$ (5) & $25.19$ & $+4.45$ (8) & $42.59$ & $+1.07$ (4) & $29.42$ & $+3.58$ (8)\\
& Independent noise (ROSE) & $42.88$ & $+0.94$ (5) & $25.96$ & $+5.22$ (8) & $42.51$ & $+0.99$ (4) & $30.24$ & $+4.40$ (8)\\
\addlinespace[3pt]
Mass $\mathcal M$ & ROSE, added mass & $42.50$ & $+0.57$ (5) & $25.80$ & $+5.06$ (8) & $41.16$ & $-0.35$ (3) & $28.37$ & $+2.53$ (5)\\
\bottomrule\end{tabular}\end{adjustbox}
\end{table}

\subsection{Sensitivity to ROSE's bandwidth}
\label{app:bandwidth}
ROSE is run everywhere with the window of Appendix~\ref{app:interface},
close to one standard deviation of the labelled anomalies in each feature.
We multiply this window by $s\in\{0.1,0.25,0.5,2\}$, with the source rows
and unit noise of $s=1$ (ROSE) and $s=0$ (ROS): default \textsc{XGBGraph}
and \textsc{RFGraph} on $\Phi$, and BWGNN with synthetic input nodes
(Appendix~\ref{app:input-rose}), on the 17 benchmarks. No bandwidth is
selected on validation; Table~\ref{tab:app-bandwidth} gives the whole
curve and, as an optimistic upper bound, the gain when the best $s$ of
each dataset is chosen in hindsight on the test AUPRC.

On the injected panel the gain of the trees does not depend on the
amplitude once there is noise at all: one tenth of the window already
gives $+4.46$ (\textsc{XGBGraph}) and $+3.73$ (\textsc{RFGraph}) where ROS
gives $-0.49$ and $+0.06$, and $s=1$ is at or near the top of both curves.
BWGNN gains at no bandwidth, on either panel (between $-1.17$ and $+0.11$,
no interval excludes zero), so its null result is not an effect of too
strong a perturbation. On the organic panel \textsc{XGBGraph} stays
between $+0.54$ and $+1.08$, unresolved. \textsc{RFGraph} does better
there with a narrower window: $+1.75$ $[+0.14,+3.58]$ at $s=0.1$, and
$s=0.25$ is above $s=1$ by $+0.85$ $[+0.37,+1.37]$; this is one
unadjusted interval among ten organic tree cells, and it stays below two
points. Even the hindsight choice of $s$ leaves the organic gains at
$+1.58$ and $+2.09$ for the trees and $+0.62$ $[-0.74,+1.65]$ for BWGNN.

\afterpage{\begin{widetables}
\begin{table}[!htbp]
\centering\small
\setlength{\tabcolsep}{3pt}
\caption{\textbf{ROSE at other bandwidths} (\Scarce{20}, AUPRC points over class weighting, 95\% interval). The window of ROSE is multiplied by $s$: $s=1$ is ROSE and $s=0$ is ROS, with the same sources and unit noise in every column. BWGNN: synthetic input nodes (Appendix~\ref{app:input-rose}), $s=0$ not run. Best: the $s$ with the highest mean test AUPRC, chosen per dataset in hindsight (an optimistic upper bound). Bold: interval excludes zero.}
\label{tab:app-bandwidth}
\begin{adjustbox}{max width=\linewidth}
\begin{tabular}{lllllllll}\toprule
Detector & Panel & $s=0$ & $s=0.1$ & $s=0.25$ & $s=0.5$ & $s=1$ & $s=2$ & Best\\\midrule
\textsc{XGBGraph} & Organic & $-0.02$ $[-0.31,+0.33]$ & $+0.93$ $[-0.05,+2.03]$ & $+0.81$ $[-0.26,+1.97]$ & $+1.08$ $[-0.06,+2.36]$ & $+0.94$ $[-0.33,+2.43]$ & $+0.54$ $[-0.44,+1.61]$ & $\mathbf{+1.58}$ $[+0.47,+2.84]$\\
 & Injected & $-0.49$ $[-1.33,+0.19]$ & $\mathbf{+4.46}$ $[+2.36,+6.36]$ & $\mathbf{+4.36}$ $[+2.07,+6.27]$ & $\mathbf{+4.92}$ $[+2.37,+7.21]$ & $\mathbf{+5.22}$ $[+2.69,+7.44]$ & $\mathbf{+4.93}$ $[+2.42,+7.07]$ & $\mathbf{+5.78}$ $[+3.63,+7.70]$\\
\addlinespace[3pt]
\textsc{RFGraph} & Organic & $+0.23$ $[-0.23,+0.74]$ & $\mathbf{+1.75}$ $[+0.14,+3.58]$ & $+1.83$ $[-0.07,+4.04]$ & $+1.45$ $[-0.56,+3.83]$ & $+0.99$ $[-1.08,+3.42]$ & $+0.45$ $[-1.57,+2.91]$ & $\mathbf{+2.09}$ $[+0.33,+4.18]$\\
 & Injected & $+0.06$ $[-1.02,+1.17]$ & $\mathbf{+3.73}$ $[+0.46,+7.66]$ & $\mathbf{+3.97}$ $[+0.52,+7.91]$ & $\mathbf{+4.36}$ $[+0.80,+8.35]$ & $\mathbf{+4.40}$ $[+0.48,+8.53]$ & $+3.40$ $[-0.26,+7.18]$ & $\mathbf{+5.70}$ $[+2.71,+9.27]$\\
\addlinespace[3pt]
BWGNN & Organic & --- & $+0.11$ $[-1.16,+1.17]$ & $+0.08$ $[-1.30,+0.98]$ & $-0.74$ $[-2.99,+0.72]$ & $-1.17$ $[-3.50,+0.50]$ & $-0.37$ $[-2.54,+1.17]$ & $+0.62$ $[-0.74,+1.65]$\\
 & Injected & --- & $+0.01$ $[-0.82,+0.76]$ & $-0.46$ $[-1.90,+0.61]$ & $+0.06$ $[-0.67,+0.73]$ & $+0.06$ $[-1.17,+1.08]$ & $-0.13$ $[-2.83,+2.02]$ & $\mathbf{+1.27}$ $[+0.13,+2.43]$\\
\bottomrule\end{tabular}\end{adjustbox}
\end{table}

\end{widetables}}

\subsection{Multivariate Gaussian SMOTE}
\label{app:mgs}
MGS \citep{sakho2026rebalancing} draws a minority point uniformly, takes
it with its $K$ nearest minority neighbours, and samples from the
Gaussian that has the mean of these $K+1$ points and their covariance,
normalised by $K+1$. We ran it in the default trees, TabPFN on $\Phi$ and
the four GNNs: 60 rows to parity, unit weights, and the splits and
placement of the other samplers (rows of $\Phi$; hidden representation
for the GNNs).
The published default is $K=d+1$, and the authors' package
(\texttt{mgs-grf}~0.1.0) uses $K=d$. $\Phi$ has between 30 and 36{,}141
columns, so both exceed the 19 neighbours that 20 anomalies allow, and the
package raises an error. Before reading any result we fixed $K=19$, the
largest admissible value, as the primary arm, and $K=5$, the
neighbourhood of our SMOTE, as a sensitivity arm. At $K=19$ every
neighbourhood is the whole labelled set, so MGS reduces to one Gaussian
fitted to the 20 anomalies, of rank at most 19. The package forms each
$d\times d$ covariance, which is not feasible on the widest tables; we
draw $\hat\mu+C^{\top}z/\sqrt{K+1}$ instead, with $C$ the $(K+1)\times d$
matrix of centred neighbours and $z\sim\mathcal N(0,I_{K+1})$, whose
covariance $C^{\top}C/(K+1)$ is exactly that of MGS (without the
package's $10^{-8}I$ term). On one split of T-Finance (30 columns),
200{,}000 draws from the package and from our sampler both reproduce the
mean and covariance of the MGS mixture at both values of $K$, to within
$0.4\%$ of a standard deviation and $2\%$ of the product of standard
deviations. In the GNNs the same draw is taken on the hidden
representation at every epoch; $\hat\mu$ and $C$ keep their gradient, so
the synthetic rows back-propagate to the labelled anomalies as for the
other samplers. In SpaceGNN the draw is made within each mini-batch,
whose anomalies cap $K$.

Table~\ref{tab:app-mgs} gives the contrasts; the gains and losses named
here are resolved. In the default trees, MGS
at $K=19$ gains over class weighting on the injected panel
and on attribute swaps, not on the organic panel or on cliques, and does
not differ from ROSE in any contrast; at $K=5$ it is below ROSE on
swaps. In TabPFN it gains at $K=19$ on the injected panel and
on swaps, where it is $1.9$ points below ROSE; at $K=5$ it gains on
neither panel. In the four GNNs it neither gains nor loses over class
weighting on either panel or on swaps, at either $K$.

\begin{table}[!htbp]
\centering\small
\setlength{\tabcolsep}{4pt}
\caption{\textbf{Multivariate Gaussian SMOTE} (\Scarce{20}, paired differences in AUPRC points, 95\% interval). $K=19$: primary arm, one Gaussian fitted to the 20 labelled anomalies; $K=5$: sensitivity arm; in SpaceGNN, $K$ is capped by the anomalies of each mini-batch. Ref.: class weighting (unresampled context for TabPFN). Swaps and cliques: the eight injected benchmarks with type labels. TFM4GAD and the tuned trees were not run.}
\label{tab:app-mgs}
\begin{adjustbox}{max width=\linewidth}
\begin{tabular}{lllllll}\toprule
Detector & $K$ & MGS minus & Organic (8) & Injected (9) & Swaps (8) & Cliques (8)\\\midrule
TabPFN on $\Phi$ & $19$ & Ref. & $-0.78$ $[-2.38,+0.55]$ & $\mathbf{+2.24}$ $[+1.04,+3.44]$ & $\mathbf{+1.66}$ $[+0.40,+3.14]$ & $+1.10$ $[-0.39,+2.76]$\\
 &  & ROSE & $-0.53$ $[-1.33,+0.13]$ & $-0.28$ $[-1.43,+0.77]$ & $\mathbf{-1.94}$ $[-3.92,-0.48]$ & $\mathbf{+0.85}$ $[+0.14,+1.65]$\\
 &  & SMOTE & $+0.61$ $[-0.45,+2.15]$ & $+1.05$ $[-0.56,+2.60]$ & $\mathbf{+4.28}$ $[+1.29,+7.75]$ & $-1.91$ $[-4.59,+0.49]$\\
\addlinespace[3pt]
 & $5$ & Ref. & $-0.64$ $[-2.16,+0.70]$ & $+1.12$ $[-0.02,+2.24]$ & $-1.11$ $[-2.97,+0.24]$ & $\mathbf{+2.19}$ $[+0.40,+4.16]$\\
 &  & ROSE & $-0.39$ $[-1.27,+0.34]$ & $-1.40$ $[-3.47,+0.32]$ & $\mathbf{-4.71}$ $[-8.41,-1.59]$ & $\mathbf{+1.95}$ $[+0.36,+4.10]$\\
 &  & SMOTE & $\mathbf{+0.76}$ $[+0.04,+1.60]$ & $-0.06$ $[-1.04,+0.88]$ & $+1.51$ $[-0.59,+4.22]$ & $-0.81$ $[-2.29,+0.66]$\\
\addlinespace[3pt]
\textsc{RFGraph}, default & $19$ & Ref. & $+0.79$ $[-2.01,+3.95]$ & $\mathbf{+4.80}$ $[+0.88,+8.93]$ & $\mathbf{+10.96}$ $[+3.15,+20.34]$ & $-2.84$ $[-8.13,+1.90]$\\
 &  & ROSE & $-0.20$ $[-1.28,+0.93]$ & $+0.40$ $[-0.22,+1.07]$ & $+0.19$ $[-0.73,+1.27]$ & $+0.71$ $[-0.08,+1.71]$\\
 &  & SMOTE & $+0.27$ $[-1.50,+2.22]$ & $+3.04$ $[-0.52,+6.71]$ & $\mathbf{+8.84}$ $[+2.41,+16.34]$ & $-3.39$ $[-8.28,+0.21]$\\
\addlinespace[3pt]
 & $5$ & Ref. & $+1.40$ $[-0.53,+3.75]$ & $\mathbf{+3.48}$ $[+0.97,+5.88]$ & $\mathbf{+6.30}$ $[+1.34,+12.70]$ & $-0.71$ $[-5.03,+3.96]$\\
 &  & ROSE & $+0.41$ $[-0.39,+1.16]$ & $-0.92$ $[-3.25,+1.14]$ & $\mathbf{-4.47}$ $[-8.35,-1.37]$ & $\mathbf{+2.83}$ $[+0.70,+5.50]$\\
 &  & SMOTE & $+0.88$ $[-0.13,+2.07]$ & $+1.72$ $[-0.12,+3.38]$ & $\mathbf{+4.18}$ $[+0.77,+8.63]$ & $-1.27$ $[-4.40,+1.36]$\\
\addlinespace[3pt]
\textsc{XGBGraph}, default & $19$ & Ref. & $+0.06$ $[-2.56,+2.45]$ & $\mathbf{+4.68}$ $[+2.24,+6.90]$ & $\mathbf{+5.41}$ $[+1.48,+10.56]$ & $+3.33$ $[-2.02,+8.58]$\\
 &  & ROSE & $-0.88$ $[-3.01,+0.59]$ & $-0.53$ $[-1.50,+0.41]$ & $-0.22$ $[-1.49,+1.05]$ & $-0.56$ $[-2.32,+0.78]$\\
 &  & SMOTE & $-0.10$ $[-1.91,+1.47]$ & $+1.24$ $[-1.01,+3.44]$ & $\mathbf{+3.69}$ $[+0.96,+7.17]$ & $-1.36$ $[-4.78,+1.80]$\\
\addlinespace[3pt]
 & $5$ & Ref. & $+0.79$ $[-0.48,+2.32]$ & $\mathbf{+5.11}$ $[+2.63,+7.43]$ & $\mathbf{+4.24}$ $[+0.83,+8.91]$ & $+5.44$ $[-0.31,+11.70]$\\
 &  & ROSE & $-0.15$ $[-1.03,+0.72]$ & $-0.11$ $[-1.35,+1.13]$ & $\mathbf{-1.40}$ $[-3.10,-0.07]$ & $+1.55$ $[-0.26,+3.57]$\\
 &  & SMOTE & $+0.63$ $[-0.08,+1.44]$ & $\mathbf{+1.66}$ $[+0.43,+2.93]$ & $\mathbf{+2.52}$ $[+0.51,+5.25]$ & $+0.75$ $[-1.29,+3.03]$\\
\addlinespace[3pt]
SpaceGNN & $19$ & Ref. & $-0.08$ $[-0.75,+0.43]$ & $-0.15$ $[-1.38,+1.03]$ & $+0.99$ $[-0.46,+2.68]$ & $-1.20$ $[-3.12,+0.20]$\\
 &  & ROSE & $+0.31$ $[-0.24,+0.87]$ & $-0.82$ $[-2.21,+0.40]$ & $+0.27$ $[-1.08,+1.77]$ & $-1.56$ $[-3.96,+0.02]$\\
 &  & SMOTE & $-0.32$ $[-1.10,+0.26]$ & $-0.56$ $[-2.18,+0.88]$ & $+0.79$ $[-0.43,+2.45]$ & $-1.77$ $[-4.82,+0.28]$\\
\addlinespace[3pt]
 & $5$ & Ref. & $+0.04$ $[-0.41,+0.57]$ & $-0.15$ $[-1.26,+0.93]$ & $+0.51$ $[-0.89,+1.97]$ & $-0.66$ $[-2.40,+0.65]$\\
 &  & ROSE & $+0.43$ $[-0.34,+1.58]$ & $-0.82$ $[-2.05,+0.34]$ & $-0.20$ $[-1.33,+1.02]$ & $-1.02$ $[-3.10,+0.46]$\\
 &  & SMOTE & $-0.20$ $[-0.65,+0.23]$ & $-0.56$ $[-2.01,+0.90]$ & $+0.31$ $[-1.39,+2.29]$ & $-1.23$ $[-3.71,+0.41]$\\
\addlinespace[3pt]
GHRN & $19$ & Ref. & $-1.34$ $[-3.79,+0.22]$ & $-0.08$ $[-1.24,+0.95]$ & $-0.41$ $[-2.58,+1.56]$ & $+0.19$ $[-0.20,+0.76]$\\
 &  & ROSE & $-0.39$ $[-1.85,+0.85]$ & $-0.10$ $[-1.50,+1.20]$ & $-0.40$ $[-2.79,+1.86]$ & $+0.05$ $[-0.60,+0.72]$\\
 &  & SMOTE & $-1.39$ $[-3.74,+0.24]$ & $-0.28$ $[-1.31,+0.74]$ & $+0.00$ $[-1.86,+2.04]$ & $-0.41$ $[-1.10,+0.11]$\\
\addlinespace[3pt]
 & $5$ & Ref. & $-0.32$ $[-1.74,+1.69]$ & $-0.06$ $[-1.18,+0.91]$ & $-1.30$ $[-3.18,+0.25]$ & $\mathbf{+0.86}$ $[+0.11,+1.93]$\\
 &  & ROSE & $+0.63$ $[-1.06,+3.13]$ & $-0.08$ $[-1.39,+1.18]$ & $-1.28$ $[-3.67,+1.03]$ & $\mathbf{+0.72}$ $[+0.05,+1.79]$\\
 &  & SMOTE & $-0.37$ $[-1.72,+0.92]$ & $-0.26$ $[-1.44,+0.89]$ & $-0.88$ $[-2.97,+1.40]$ & $+0.26$ $[-0.20,+0.89]$\\
\addlinespace[3pt]
BWGNN & $19$ & Ref. & $-0.26$ $[-1.14,+0.56]$ & $+0.18$ $[-0.63,+1.04]$ & $+0.00$ $[-1.62,+1.77]$ & $+0.25$ $[-0.12,+0.74]$\\
 &  & ROSE & $-0.10$ $[-1.33,+1.16]$ & $+0.18$ $[-0.73,+1.15]$ & $-0.11$ $[-1.63,+1.65]$ & $+0.29$ $[-0.15,+0.87]$\\
 &  & SMOTE & $-0.29$ $[-1.55,+1.08]$ & $+0.30$ $[-0.83,+1.58]$ & $+0.63$ $[-1.46,+3.03]$ & $-0.12$ $[-0.53,+0.17]$\\
\addlinespace[3pt]
 & $5$ & Ref. & $+0.47$ $[-0.61,+1.83]$ & $-0.67$ $[-2.55,+0.84]$ & $-2.25$ $[-5.62,+0.37]$ & $+0.77$ $[-0.06,+1.92]$\\
 &  & ROSE & $+0.64$ $[-0.64,+2.48]$ & $-0.67$ $[-2.03,+0.41]$ & $\mathbf{-2.36}$ $[-4.92,-0.46]$ & $\mathbf{+0.81}$ $[+0.08,+1.80]$\\
 &  & SMOTE & $+0.44$ $[-0.95,+2.53]$ & $-0.55$ $[-1.83,+0.51]$ & $-1.62$ $[-4.01,+0.47]$ & $+0.41$ $[-0.01,+0.99]$\\
\addlinespace[3pt]
GATv2 & $19$ & Ref. & $+0.04$ $[-0.88,+1.18]$ & $-0.67$ $[-1.50,+0.08]$ & $-1.55$ $[-3.45,+0.05]$ & $+0.04$ $[-0.65,+0.63]$\\
 &  & ROSE & $+0.54$ $[-0.63,+2.34]$ & $-0.42$ $[-1.66,+0.63]$ & $-1.35$ $[-3.80,+0.56]$ & $+0.43$ $[-0.66,+1.48]$\\
 &  & SMOTE & $+0.44$ $[-0.67,+2.15]$ & $-0.65$ $[-1.64,+0.23]$ & $-1.36$ $[-3.37,+0.66]$ & $+0.07$ $[-0.82,+0.79]$\\
\addlinespace[3pt]
 & $5$ & Ref. & $-0.11$ $[-1.09,+0.94]$ & $-0.04$ $[-1.56,+1.50]$ & $-1.76$ $[-4.39,+0.75]$ & $+1.17$ $[-0.05,+2.85]$\\
 &  & ROSE & $+0.40$ $[-0.96,+2.17]$ & $+0.21$ $[-1.35,+1.79]$ & $-1.56$ $[-4.23,+0.96]$ & $+1.56$ $[-0.14,+3.75]$\\
 &  & SMOTE & $+0.30$ $[-1.00,+1.90]$ & $-0.02$ $[-1.36,+1.27]$ & $-1.57$ $[-3.54,+0.22]$ & $+1.20$ $[-0.00,+2.83]$\\
\bottomrule\end{tabular}\end{adjustbox}
\end{table}

\subsection{Selecting the strategy on validation data}
For each detector and split we keep the strategy with the highest
validation AUPRC (20 anomalies and 80 normals), either between the
reference and ROSE or among the strategies of
Tables~\ref{tab:cross-family} and~\ref{tab:app-cross-family-other}
available for that detector (ROSE on $X$ and MGS excluded), and compare its test
AUPRC with that of the reference (Table~\ref{tab:app-selection}). For the
GNNs and the tuned trees the same validation set has already selected the
checkpoint or the configuration.

\afterpage{\begin{widetables}
\begin{table}[!htbp]
\centering\small
\setlength{\tabcolsep}{4pt}
\caption{\textbf{Selecting the strategy on validation data} (\Scarce{20}). Test AUPRC gain, over always using the reference, of the strategy with the best validation AUPRC in each split; 95\% interval, bold when it excludes zero. Pair: reference or ROSE; All: the $n$ strategies of Tables~\ref{tab:cross-family} and \ref{tab:app-cross-family-other} available for the detector (ROSE on $X$ and MGS excluded). Oracle: mean gain on the organic panel when each benchmark uses the strategy with the best mean test AUPRC over its ten splits; fixed per benchmark, so per-split selection can exceed it.}
\label{tab:app-selection}
\begin{adjustbox}{max width=\linewidth}
\begin{tabular}{lllrrlll}\toprule
& \multicolumn{3}{c}{Pair} & & \multicolumn{3}{c}{All}\\
\cmidrule(lr){2-4}\cmidrule(lr){6-8}
Detector & Organic & Oracle & Injected & $n$ & Organic & Oracle & Injected\\\midrule
TFM4GAD & $+0.08$ $[-0.35,+0.54]$ & $+0.24$ & $+0.60$ $[-0.67,+1.99]$ & 3 & $-0.03$ $[-0.58,+0.54]$ & $+0.52$ & $\mathbf{+2.65}$ $[+0.79,+5.25]$\\
TabPFN on $\Phi$ & $+0.24$ $[-0.44,+1.08]$ & $+0.65$ & $\mathbf{+2.22}$ $[+1.02,+3.67]$ & 6 & $+0.34$ $[-0.54,+1.30]$ & $+0.99$ & $\mathbf{+3.14}$ $[+1.42,+4.87]$\\
\textsc{RFGraph}, tuned & $-0.26$ $[-1.20,+0.51]$ & $+0.40$ & $+1.41$ $[-0.81,+3.94]$ & 5 & $-0.09$ $[-0.76,+0.60]$ & $+0.76$ & $+1.91$ $[-0.44,+4.51]$\\
\textsc{XGBGraph}, tuned & $+0.60$ $[-0.19,+1.46]$ & $+1.07$ & $\mathbf{+1.66}$ $[+0.06,+3.63]$ & 5 & $+0.61$ $[-0.14,+1.49]$ & $+1.40$ & $+1.36$ $[-0.19,+3.15]$\\
\textsc{RFGraph}, default & $+1.38$ $[-0.21,+3.52]$ & $+1.73$ & $\mathbf{+4.24}$ $[+1.17,+7.94]$ & 8 & $+1.65$ $[-0.12,+3.77]$ & $+2.25$ & $\mathbf{+4.62}$ $[+2.00,+7.71]$\\
\textsc{XGBGraph}, default & $\mathbf{+1.19}$ $[+0.03,+2.60]$ & $+1.18$ & $\mathbf{+4.71}$ $[+2.97,+6.55]$ & 8 & $\mathbf{+1.05}$ $[+0.15,+2.14]$ & $+1.52$ & $\mathbf{+4.91}$ $[+2.93,+7.02]$\\
SpaceGNN & $-0.05$ $[-0.55,+0.35]$ & $+0.08$ & $\mathbf{+0.89}$ $[+0.29,+1.59]$ & 6 & $+0.27$ $[-0.28,+0.86]$ & $+0.46$ & $\mathbf{+1.14}$ $[+0.27,+2.14]$\\
GHRN & $-0.28$ $[-1.29,+0.41]$ & $+0.06$ & $+0.52$ $[-0.04,+1.15]$ & 6 & $-0.34$ $[-1.93,+1.09]$ & $+0.34$ & $+0.54$ $[-0.49,+1.43]$\\
BWGNN & $+0.19$ $[-0.36,+0.83]$ & $+0.22$ & $+0.31$ $[-0.16,+0.89]$ & 8 & $+0.17$ $[-1.25,+1.39]$ & $+0.57$ & $+0.17$ $[-0.82,+1.04]$\\
GATv2 & $+0.87$ $[-0.04,+1.98]$ & $+0.09$ & $-0.08$ $[-0.86,+0.70]$ & 8 & $\mathbf{+2.55}$ $[+0.44,+5.48]$ & $+1.29$ & $+0.76$ $[-0.67,+2.20]$\\
\bottomrule\end{tabular}\end{adjustbox}
\end{table}

\end{widetables}}

\section{Implementation Details}
\label{app:xgb-configurations}\label{app:ff}\label{app:train}\label{sec:baselines}\label{app:tuning}\label{app:ports}
The file \texttt{IMPLEMENTATION.md} of the released repository gives the
implementation details, each with the module that implements it: the two
\textsc{XGBGraph} configurations (100 trees at learning rate 0.3; 200 trees
at 0.05 for the mechanism controls) and their refresh schedules; the
parameters of the structural views and of the residual-guided controls;
the settings of ADASYN, SMOTE and GODM; the random search of the tuned
trees and its deviations from GADBench; and, for every reference detector,
its source, pinned upstream commit, configuration and deviations from the
upstream code. Three facts bear on the comparisons. Each arm of the tuned
trees selects its own candidate, among the same 100, on the validation
AUPRC of its split. Only the tree detectors are tuned: the GNNs and TFM4GAD
run at the configuration their authors ship for the dataset or, where
there is none, at a transfer configuration fixed before their results were
read. And each detector keeps its own correction for imbalance.

\FloatBarrier
\section{Experiment Details}
\label{sec:exp}

\subsection{Datasets}
\label{sec:datasets}
Table~\ref{tab:datasets} separates eight datasets with observed anomaly
labels from nine injected benchmarks, which follow the DOMINANT/CoLA
recipe \citep{dominant,cola}: half of the anomalies are structural
(cliques) and half contextual (attributes replaced by those of a distant
node in feature space). The six legacy supports use the anomaly files
distributed with CoLA. We injected the three additional graphs ourselves
with the standard parameters: cliques of 15 nodes, 50 candidates for each
attribute swap, anomalies on $5\%$ of the nodes (half of them structural,
rounded down to whole cliques, and the remainder contextual), one
realisation, seed 3407. Type labels ship with the CoLA files for Cora,
Citeseer, Pubmed, BlogCatalog and Flickr and were rebuilt for our three
graphs by replaying the injection seed; in these eight datasets the two
types partition the anomaly labels exactly. ACM ships no type labels and
is excluded from by-type analyses.

\begin{table}[!htbp]
    \centering
    \small
    \setlength{\tabcolsep}{5pt}
    \caption{Statistics of the $17$-dataset panel, recomputed from the loaded
    DGL graphs. \emph{Edges}: directed edges, self-loops removed;
    \emph{Avg.\ deg.}: mean out-degree; \emph{Homoph.}: fraction of edges
    whose endpoints share a label.}
    \label{tab:datasets}
    \begin{tabular}{lrrrrrr}
        \toprule
        Dataset & Nodes & Edges & Feat. & Anom.\ \% & Avg.\ deg. & Homoph. \\
        \midrule
        \multicolumn{7}{l}{\emph{Organic anomalies (real labels)}} \\
        \midrule
        Amazon       &    11,944 &   8,835,152 &     25 &  6.9 &  739.7 & 0.95 \\
        FDCompCN     &     5,317 &       7,268 &     57 & 10.5 &    1.4 & 0.85 \\
        Questions    &    48,921 &     153,540 &    301 &  3.0 &    3.1 & 0.84 \\
        Reddit       &    10,984 &     157,032 &     64 &  3.3 &   14.3 & 0.95 \\
        T-Finance    &    39,357 &  42,445,086 &     10 &  4.6 & 1078.5 & 0.97 \\
        Tolokers     &    11,758 &     519,000 &     10 & 21.8 &   44.1 & 0.59 \\
        Weibo        &     8,405 &     407,963 &    400 & 10.3 &   48.5 & 0.98 \\
        YelpChi      &    45,954 &   7,693,958 &     32 & 14.5 &  167.4 & 0.77 \\
        \midrule
        \multicolumn{7}{l}{\emph{Injected anomalies, legacy supports}} \\
        \midrule
        ACM              & 16,484 & 148,146 &  8,337 & 3.6 &  9.0 & 0.93 \\
        BlogCatalog      &  5,196 & 345,566 &  8,189 & 5.8 & 66.5 & 0.89 \\
        Citeseer         &  3,327 &  10,154 &  3,703 & 4.5 &  3.1 & 0.93 \\
        Cora             &  2,708 &  11,606 &  1,433 & 5.5 &  4.3 & 0.91 \\
        Flickr           &  7,575 & 482,608 & 12,047 & 5.9 & 63.7 & 0.90 \\
        Pubmed           & 19,717 &  92,846 &    500 & 3.0 &  4.7 & 0.94 \\
        \midrule
        \multicolumn{7}{l}{\emph{Injected anomalies, additional benchmark graphs}} \\
        \midrule
        Amazon-Computers & 13,752 & 496,332 &    767 & 5.0 & 36.1 & 0.91 \\
        Amazon-Photo     &  7,650 & 240,672 &    745 & 5.0 & 31.5 & 0.91 \\
        Coauthor-CS      & 18,333 & 170,084 &  6,805 & 5.0 &  9.3 & 0.90 \\
        \bottomrule
    \end{tabular}
\end{table}

\subsection{Evaluation Protocol and Reproducibility}
\label{sec:protocol}

Evaluation is transductive: methods access the complete graph and attributes,
but only training labels for fitting. Methods share ten paired splits
(seeds $3407+10\cdot\text{trial}$, $\text{trial}\in\{0,\ldots,9\}$).
Within each resampling contrast, canonical features and learner settings
are fixed.

\paragraph{Label budgets and evaluation nodes.}
\Scarce{20} uses twenty labelled anomalies and eighty labelled normal nodes,
with a disjoint validation set of the same class proportions.
The label-budget grid (Table~\ref{tab:app-grid}) crosses $k\in\{5,10,20\}$
labelled anomalies with $n\in\{20,40,80\}$ labelled normals on Cora,
Flickr, Coauthor-CS, Weibo, FDCompCN and T-Finance. Training labels are
nested in $k$ at fixed $n$ and in $n$ at fixed $k$; validation (20
anomalies and 80 normals) and test nodes are identical across the nine
cells, and unused training anomalies never return to evaluation. ROSE
adds $n-k$ synthetic rows, so its dose moves with the ratio; in the
$(20,20)$ cell the classes are balanced, so class weighting and ROSE
reduce to unweighted training. Runs on the official fully supervised
masks are in Appendix~\ref{app:fullsup}.

\paragraph{Selection and exploratory status.}
Only the first default-tree comparison (class weighting, ROS, ROSE,
SMOTE-$\Phi$ and island views with both learners, GODM with
\textsc{XGBGraph}; Tables~\ref{tab:cross-family}
and~\ref{tab:app-cross-family-other}) was fixed before its test
results were read: its configurations were selected on validation data,
and its candidates, comparisons and analysis were set beforehand, without
external preregistration. Every other experiment is exploratory and was
added afterwards, on the same cached splits: ADASYN, GODM with
\textsc{RFGraph}, the
residual-guided reruns and matched controls with both tree learners, the
tuned trees, the GNN detectors and TFM4GAD, TabPFN on $\Phi$, the samplers
inside the four GNNs at the hidden layer and at the input, frozen
encoders, the by-type analyses, the label-free scorers and columns, the
block-restricted and symmetric ROSE arms, the same-graph injection, the
two recipe variants, the $(k,n)$ grid, the selection on validation
data and MGS. No model, configuration or checkpoint was selected on test data,
but these experiments were chosen after test results were known.
Intervals are not adjusted for multiplicity.

\paragraph{Reproduction.}
The environment is Python~3.10.20, numpy~2.2.6, scikit-learn~1.7.2,
xgboost~1.7.6 and imbalanced-learn~0.14.2, with PyTorch~2.7.0 (CUDA~12.8)
and DGL~2.4.0 for the GNNs and PyTorch Geometric~2.6.1 for GODM. The
in-context models use tabpfn~6.0.6, loaded explicitly as model version~2
with the classifier checkpoint \texttt{tabpfn-v2-classifier.ckpt}
(Hugging Face repository \texttt{Prior-Labs/TabPFN-v2-clf}). All
experiments ran on one server with an NVIDIA RTX PRO 6000 Blackwell GPU
(96\,GB), two 16-core AMD EPYC 9135 processors and 1\,TB of memory. The
per-trial logs record about 300 hours of fitting time in total, summed
over fits that often ran concurrently; the random search of the tuned
trees (about 85 hours) and ConsisGAD (about 70 hours) account for half
of it. The released repository contains
configurations, run manifests and per-trial results (dataset, split,
label budget, strategy, fallback, validation and test metrics).

\subsection{Metrics and Statistical Analysis}
\label{sec:statistics}
AUPRC is the primary metric; AUROC and Recall@$K$ ($K$ is the test-set
anomaly count) are reported for the leaderboard, the anchors and the
primary diagnostic contrasts; their per-dataset values and the Brier
score are in the released repository. The paired difference is
$\Delta=\mathrm{Metric}_{\mathrm{candidate}}-\mathrm{Metric}_{\mathrm{reference}}$.
We average seeds within each dataset, then datasets equally within the
stated panel, and count datasets with positive mean differences as wins.
AUPRC scores and differences are multiplied by 100. Intervals are 95\%
hierarchical-bootstrap intervals (20,000 replicates, resampling datasets
and then paired seeds); they quantify estimation uncertainty rather than
equivalence. The bootstrap samples depend on the order in which a script
lists the paired trials, so the limits of the same contrast can differ by
up to about 0.1 between tables.

\paragraph{Descriptive variance decomposition.}
For each organic dataset and learner, let $x_{ij}$ denote \Scarce{20} test
AUPRC for split $i\in\{1,\ldots,10\}$ and strategy
$j\in\{1,\ldots,5\}$: class weighting, ROS, ROSE, SMOTE-$\Phi$, and
uniform island views. The complete, unreplicated $10\times5$ layout gives
\[
\underbrace{\textstyle\sum_{ij}(x_{ij}-\bar x)^2}_{SS_{\mathrm{tot}}}
=
\underbrace{5\textstyle\sum_i(\bar x_{i\cdot}-\bar x)^2}_{SS_{\mathrm{split}}}
+
\underbrace{10\textstyle\sum_j(\bar x_{\cdot j}-\bar x)^2}_{SS_{\mathrm{strategy}}}
+SS_{\mathrm{resid}}.
\]
The residual pools interaction and noise, which one observation per cell
cannot separate. Shares $SS_\bullet/SS_{\mathrm{tot}}$ are computed
within each dataset and averaged equally across the eight organic
datasets; the decomposition is descriptive, not an inferential test.
Split and strategy shares are $72/10\%$ for XGBGraph and $68/16\%$ for
RFGraph; Weibo is the only dataset where the strategy share exceeds the
split share ($35$ against $29\%$, and $47$ against $27\%$). The split
share remains the larger in three variants: dropping FDCompCN ($69/11\%$
and $66/18\%$), using AUROC ($73/8\%$ and $66/18\%$), and adding
unweighted training as a sixth strategy ($72/9\%$ and $65/18\%$).
Table~\ref{tab:app-variance-detectors} repeats the decomposition for all
ten configurations, with the strategies available for each detector among
its reference, unweighted training, ROS, ROSE and SMOTE. Without island
views and with unweighted training, the default-tree shares become
$74/6\%$ and $67/16\%$, and on Weibo the strategy share of
\textsc{XGBGraph} falls from $35\%$ to $9\%$, below its split share
($53\%$).

\FloatBarrier
\subsection{Detector leaderboard}
\label{app:tables}

In the tables of this appendix, \emph{No resampling} (NR) is unweighted
training, SMOTE-$\Phi$ and ADASYN-$\Phi$ are SMOTE and ADASYN applied to
the rows of $\Phi$, contextual and structural anomalies are the attribute
swaps and cliques of the main text, and seeds are the ten labelled splits.
Tables~\ref{tab:app-leaderboard-org} and~\ref{tab:app-leaderboard-inj}
give the per-dataset detector leaderboard; per-dataset AUPRC, AUROC and Recall@$K$ of every strategy
with the default trees are in the released repository. The table
omits four detectors that we also ran on these splits: LightGAD+
\citep{lightgad}, a balanced random forest \citep{balancedrf} and an
isolation forest on $\Phi$, and APF \citep{apf}. Our port of APF reaches a
mean AUPRC of 41.4 on the seven GADBench datasets of the panel, against
49.6 in its paper, so we do not use it as a reference. Their results are
in the released repository.

\afterpage{\begin{widetables}
\begin{table}[!htbp]
\centering
\small
\setlength{\tabcolsep}{3pt}
\caption{\textbf{Detector leaderboard, organic panel} (\Scarce{20}, test AUPRC $\times100$ per dataset, mean over ten seeds). Last columns: panel means of AUPRC, AUROC and Recall@$K$. The methods are ordered by their mean AUPRC over all 17 datasets (last column of Table~\ref{tab:app-leaderboard-inj}). Bold: best trained method per dataset. TabPFN on $\Phi$ (45.4 and 29.1 on the two panels, 36.7 over the 17 datasets) and GATv2 (23.0, 18.3 and 20.5) are in Table~\ref{tab:cross-family}.}
\label{tab:app-leaderboard-org}
\begin{adjustbox}{max width=\linewidth}
\begin{tabular}{lrrrrrrrrcrrr}
\toprule
& \multicolumn{8}{c}{AUPRC} & & \multicolumn{3}{c}{Panel mean} \\
\cmidrule(lr){2-9}\cmidrule(lr){11-13}
Method & Amazon & FDCompCN & Questions & Reddit & T-Fin. & Tolokers & Weibo & YelpChi & & AUPRC & AUROC & Rec@$K$ \\
\midrule
TFM4GAD & \textbf{87.09} & 23.45 & 6.23 & 4.17 & 82.15 & \textbf{45.40} & \textbf{93.57} & 28.24 &  & 46.29 & 78.73 & 45.73 \\
\textsc{RFGraph} tuned + ROSE & 84.52 & \textbf{29.99} & 9.16 & 4.19 & 78.91 & 33.81 & 85.46 & 28.01 &  & 44.26 & 77.97 & 44.48 \\
\textsc{RFGraph} tuned + CW & 85.26 & 29.00 & 8.37 & 4.18 & 81.03 & 36.01 & 87.71 & 26.64 &  & 44.78 & 78.11 & 44.65 \\
\textsc{RFGraph} default + ROSE & 70.96 & 26.86 & 8.64 & 3.84 & 79.45 & 35.15 & 87.86 & 27.28 &  & 42.51 & 76.38 & 41.78 \\
\textsc{XGBGraph} tuned + ROSE & 85.93 & 28.58 & 8.15 & 4.09 & 78.45 & 34.32 & 87.17 & \textbf{30.44} &  & 44.64 & 77.97 & 44.59 \\
\textsc{XGBGraph} tuned + CW & 85.15 & 27.37 & 6.95 & 4.37 & 77.71 & 35.33 & 85.19 & 27.77 &  & 43.73 & 77.67 & 43.59 \\
\textsc{RFGraph} default + CW & 69.95 & 25.99 & \textbf{10.03} & 4.08 & \textbf{82.29} & 36.61 & 80.87 & 22.32 &  & 41.52 & 77.23 & 40.70 \\
\textsc{XGBGraph} default + CW & 83.32 & 24.27 & 7.13 & 3.81 & 78.46 & 36.08 & 78.87 & 23.56 &  & 41.94 & 76.27 & 41.31 \\
BWGNN & 80.67 & 10.91 & 6.58 & \textbf{4.65} & 63.76 & 35.86 & 76.78 & 23.87 &  & 37.89 & 73.54 & 38.94 \\
GHRN & 81.85 & 10.79 & 6.35 & 4.63 & 65.64 & 34.69 & 75.52 & 23.71 &  & 37.90 & 72.92 & 38.58 \\
SpaceGNN & 81.71 & 11.89 & 7.02 & 4.01 & 80.13 & 34.05 & 78.84 & 23.27 &  & 40.11 & 74.22 & 39.42 \\
ConsisGAD & 82.10 & 11.05 & 8.03 & 3.54 & 80.67 & 34.07 & 57.99 & 25.04 &  & 37.81 & 71.89 & 37.62 \\
PC-GNN & 81.70 & 10.12 & 6.79 & 3.13 & 49.91 & 31.98 & 72.08 & 23.66 &  & 34.92 & 70.16 & 35.82 \\
GraphSAGE & 48.50 & 10.90 & 6.70 & 3.91 & 10.23 & 33.98 & 57.08 & 20.39 &  & 23.96 & 67.21 & 26.78 \\
\midrule
\multicolumn{13}{l}{\emph{Label-free scores (no training)}} \\
$\lVert x\rVert$ & 7.14 & 9.73 & 3.57 & 2.96 & 2.41 & 21.35 & 80.41 & 12.23 &  & 17.48 & 49.08 & 16.68 \\
Degree & 5.33 & 9.19 & 5.30 & 3.07 & 5.23 & 28.48 & 10.55 & 13.87 &  & 10.13 & 50.69 & 8.24 \\
\bottomrule
\end{tabular}
\end{adjustbox}
\vspace{2pt}

{\footnotesize
CW: class weighting. Tuned trees use the random search of Appendix~\ref{app:tuning}; GNNs are untuned (published or declared transfer configuration) and keep their own training objectives.
GHRN and SpaceGNN rows are the detector runs of Appendix~\ref{app:ports}; the reference columns of Table~\ref{tab:cross-family} come from the separate sampler runs (organic and injected panel means: GHRN 38.05 and 22.41; SpaceGNN 40.07 and 16.69, with a class-weighted loss that the published SpaceGNN does not use), within 0.22 points of these rows.
}

\end{table}

\begin{table}[!htbp]
\centering
\small
\setlength{\tabcolsep}{3pt}
\caption{\textbf{Detector leaderboard, injected panel.} Same conventions and notes as Table~\ref{tab:app-leaderboard-org}. Last column: mean AUPRC over all 17 datasets, which orders the methods.}
\label{tab:app-leaderboard-inj}
\begin{adjustbox}{max width=\linewidth}
\begin{tabular}{lrrrrrrrrrcrrrr}
\toprule
& \multicolumn{9}{c}{AUPRC} & & \multicolumn{3}{c}{Panel mean} & \multicolumn{1}{c}{AUPRC} \\
\cmidrule(lr){2-10}\cmidrule(lr){12-14}\cmidrule(lr){15-15}
Method & ACM & BlogC. & Citeseer & Cora & Flickr & Pubmed & A-Comp. & A-Photo & Co-CS & & AUPRC & AUROC & Rec@$K$ & All 17 \\
\midrule
TFM4GAD & 15.81 & 21.98 & 51.15 & 51.74 & 19.86 & 24.04 & 35.48 & 43.60 & 28.74 &  & 32.49 & 77.90 & 34.02 & 38.98 \\
\textsc{RFGraph} tuned + ROSE & \textbf{20.44} & 22.83 & 52.30 & 50.09 & 17.97 & \textbf{37.03} & 23.87 & 30.13 & 24.24 &  & 30.99 & 76.75 & 34.70 & 37.23 \\
\textsc{RFGraph} tuned + CW & 20.27 & 14.01 & \textbf{52.73} & \textbf{54.89} & 17.87 & 31.55 & 21.85 & 23.57 & 27.41 &  & 29.35 & 75.82 & 32.86 & 36.61 \\
\textsc{RFGraph} default + ROSE & 20.10 & 22.29 & 50.87 & 51.19 & 18.68 & 34.91 & 21.25 & 31.40 & 21.45 &  & 30.24 & 74.67 & 33.08 & 36.01 \\
\textsc{XGBGraph} tuned + ROSE & 13.95 & 18.48 & 48.55 & 49.55 & 15.89 & 29.30 & 20.42 & 30.04 & 22.41 &  & 27.62 & 75.17 & 32.30 & 35.63 \\
\textsc{XGBGraph} tuned + CW & 13.61 & 11.14 & 45.83 & 44.10 & 14.14 & 29.51 & 19.19 & 23.32 & 24.63 &  & 25.05 & 73.92 & 29.74 & 33.84 \\
\textsc{RFGraph} default + CW & 16.73 & 6.90 & 50.63 & 46.69 & 15.84 & 34.54 & 13.37 & 20.12 & 27.73 &  & 25.84 & 72.50 & 28.39 & 33.22 \\
\textsc{XGBGraph} default + CW & 9.41 & 6.69 & 42.16 & 42.22 & 11.07 & 20.89 & 12.83 & 18.61 & 22.80 &  & 20.74 & 69.89 & 24.02 & 30.72 \\
BWGNN & 9.83 & 28.17 & 9.31 & 10.62 & 30.34 & 5.34 & \textbf{36.36} & \textbf{44.70} & 30.21 &  & 22.76 & 69.58 & 27.63 & 29.88 \\
GHRN & 10.49 & 28.17 & 9.32 & 9.59 & 30.44 & 5.38 & 35.89 & 43.18 & 28.73 &  & 22.35 & 68.80 & 26.64 & 29.67 \\
SpaceGNN & 12.47 & 20.77 & 19.48 & 21.41 & 20.18 & 12.18 & 15.64 & 19.42 & 10.59 &  & 16.91 & 70.56 & 23.69 & 27.83 \\
ConsisGAD & 12.90 & 15.85 & 14.48 & 14.78 & 11.04 & 10.65 & 22.98 & 26.43 & 8.46 &  & 15.29 & 71.83 & 20.19 & 25.89 \\
PC-GNN & 6.08 & 24.02 & 5.01 & 5.48 & 20.59 & 5.56 & 22.56 & 19.25 & \textbf{32.13} &  & 15.63 & 60.26 & 21.00 & 24.71 \\
GraphSAGE & 10.60 & \textbf{28.74} & 14.57 & 13.57 & \textbf{32.34} & 4.78 & 26.27 & 30.66 & 25.15 &  & 20.74 & 68.77 & 27.46 & 22.26 \\
\midrule
\multicolumn{15}{l}{\emph{Label-free scores (no training)}} \\
$\lVert x\rVert$ & 25.51 & 33.13 & 16.33 & 15.71 & 37.60 & 21.58 & 21.56 & 24.46 & 34.59 &  & 25.61 & 71.85 & 36.52 & 21.78 \\
Degree & 9.06 & 5.47 & 24.07 & 23.48 & 5.84 & 8.66 & 5.57 & 5.36 & 11.35 &  & 10.98 & 65.89 & 15.83 & 10.58 \\
\bottomrule
\end{tabular}
\end{adjustbox}
\end{table}
\end{widetables}}

\FloatBarrier
\subsection{Label budget}
\label{app:budget}
Table~\ref{tab:app-grid} gives the grid of Appendix~\ref{sec:protocol}.
At a fixed 1:4 ratio, four times more labels do not change ROSE's gain
over class weighting ($-2.31$ $[-8.82,+3.26]$ for \textsc{XGBGraph},
$+0.51$ $[-2.99,+3.98]$ for \textsc{RFGraph}), the contrast designated as
primary before running the grid. At fixed $n$, the gain falls as anomalies
are added on the three injected benchmarks (resolved in three of four
comparisons) but in no comparison on the three organic ones. The grid
therefore does not separate label count from class ratio. Comparisons
between cells, AUROC, Recall@$K$ and an earlier sweep over
$k\in\{5,10,20,50\}$ are in the released repository.

\begin{table}[!htbp]
\centering
\small
\setlength{\tabcolsep}{4pt}
\caption{\textbf{Label count versus class ratio: the $(k, n)$ grid} ($k$ labelled
anomalies, $n$ labelled normals; test AUPRC $\times100$). \emph{CW}: absolute
AUPRC of the class-weighted reference; other columns: paired differences,
bold when the 95\% interval excludes zero. Organic: Weibo, FDCompCN,
T-Finance; injected: Cora, Flickr, Coauthor-CS. $^{\dagger}$At $(20,20)$
the classes are balanced, so class weighting and ROSE reduce to
unweighted training.}
\label{tab:app-grid}
\begin{adjustbox}{max width=\linewidth}
\begin{tabular}{llccccccccc}
\toprule
& & \multicolumn{3}{c}{CW (absolute)} & \multicolumn{3}{c}{ROSE $-$ CW}
& \multicolumn{3}{c}{ROSE $-$ unweighted} \\
\cmidrule(lr){3-5}\cmidrule(lr){6-8}\cmidrule(lr){9-11}
Panel & $k$ & $n{=}20$ & $n{=}40$ & $n{=}80$ & $n{=}20$ & $n{=}40$ & $n{=}80$
& $n{=}20$ & $n{=}40$ & $n{=}80$ \\
\midrule
\multicolumn{11}{l}{\emph{\textsc{XGBGraph}}}\\
\midrule
Organic (3) & 5 & $27.00$ & $37.62$ & $44.40$ & $+4.01$ & $+4.75$ & $+5.39$ & $-8.54$ & $-3.43$ & $-0.77$ \\
 & 10 & $46.27$ & $49.99$ & $55.52$ & $+0.07$ & $+3.58$ & $+2.22$ & $+0.27$ & $+2.39$ & $+0.15$ \\
 & 20 & $50.36$ & $55.61$ & $60.53$ & $0^{\dagger}$ & $+1.33$ & $+1.67$ & $0^{\dagger}$ & $+1.50$ & $+1.28$ \\
\addlinespace
Injected (3) & 5 & $9.22$ & $11.72$ & $12.18$ & $\mathbf{+4.48}$ & $\mathbf{+4.49}$ & $\mathbf{+5.15}$ & $\mathbf{+2.99}$ & $\mathbf{+4.41}$ & $\mathbf{+4.46}$ \\
 & 10 & $15.39$ & $16.83$ & $18.39$ & $+2.46$ & $\mathbf{+4.06}$ & $\mathbf{+5.55}$ & $+3.36$ & $\mathbf{+4.48}$ & $\mathbf{+4.99}$ \\
 & 20 & $17.46$ & $21.53$ & $25.36$ & $0^{\dagger}$ & $+1.14$ & $+2.19$ & $0^{\dagger}$ & $+0.93$ & $+2.24$ \\
\addlinespace
All six & 5 & $18.11$ & $24.67$ & $28.29$ & $+4.24$ & $\mathbf{+4.62}$ & $\mathbf{+5.27}$ & $-2.77$ & $+0.49$ & $+1.84$ \\
 & 10 & $30.83$ & $33.41$ & $36.96$ & $+1.27$ & $\mathbf{+3.82}$ & $\mathbf{+3.89}$ & $+1.81$ & $\mathbf{+3.43}$ & $\mathbf{+2.57}$ \\
 & 20 & $33.91$ & $38.57$ & $42.95$ & $0^{\dagger}$ & $+1.23$ & $+1.93$ & $0^{\dagger}$ & $+1.21$ & $+1.76$ \\
\midrule
\multicolumn{11}{l}{\emph{\textsc{RFGraph}}}\\
\midrule
Organic (3) & 5 & $51.62$ & $53.05$ & $55.55$ & $-0.41$ & $+0.62$ & $+0.93$ & $-1.69$ & $\mathbf{-2.60}$ & $-1.28$ \\
 & 10 & $53.43$ & $55.01$ & $58.87$ & $+1.31$ & $+2.50$ & $+2.29$ & $-0.18$ & $-1.59$ & $-0.34$ \\
 & 20 & $56.45$ & $58.70$ & $63.05$ & $0^{\dagger}$ & $+3.07$ & $+1.67$ & $0^{\dagger}$ & $+0.17$ & $\mathbf{-1.34}$ \\
\addlinespace
Injected (3) & 5 & $17.70$ & $17.62$ & $16.77$ & $\mathbf{+1.40}$ & $\mathbf{+3.02}$ & $\mathbf{+5.84}$ & $\mathbf{+1.49}$ & $\mathbf{+2.75}$ & $\mathbf{+4.53}$ \\
 & 10 & $21.63$ & $23.97$ & $24.88$ & $+1.49$ & $+1.49$ & $+2.67$ & $\mathbf{+1.29}$ & $\mathbf{+1.77}$ & $\mathbf{+2.45}$ \\
 & 20 & $24.04$ & $28.41$ & $30.09$ & $0^{\dagger}$ & $-1.59$ & $+0.35$ & $0^{\dagger}$ & $-0.56$ & $+0.99$ \\
\addlinespace
All six & 5 & $34.66$ & $35.33$ & $36.16$ & $+0.50$ & $+1.82$ & $\mathbf{+3.38}$ & $-0.10$ & $+0.08$ & $+1.62$ \\
 & 10 & $37.53$ & $39.49$ & $41.88$ & $+1.40$ & $+1.99$ & $+2.48$ & $+0.55$ & $+0.09$ & $+1.05$ \\
 & 20 & $40.25$ & $43.56$ & $46.57$ & $0^{\dagger}$ & $+0.74$ & $+1.01$ & $0^{\dagger}$ & $-0.19$ & $-0.18$ \\
\bottomrule
\end{tabular}
\end{adjustbox}
\end{table}

\FloatBarrier
\subsection{Full supervision}
\label{app:fullsup}
Tables~\ref{tab:app-fullsup-xgb} and~\ref{tab:app-fullsup-rf} give the
default trees on GADBench's official fully supervised masks, with ten
seeds. Their test nodes differ from those of \Scarce{20}, so we report
means without intervals. For Amazon and YelpChi the official mask columns
repeat one split, so XGBoost without resampling has no seed variation.
ROSE's gain over class weighting is $+0.64$
(\textsc{XGBGraph}) and $-0.01$ (\textsc{RFGraph}) on the organic panel and
$+2.81$ and $+4.07$ on the injected one; against unweighted training it
is $-0.17$ and $-0.54$, then $+0.86$ and $-0.37$. With full supervision,
most of ROSE's injected gain is thus a gain over a weak reference.

\begin{table}[!htbp]
\centering
\small
\setlength{\tabcolsep}{4pt}
\caption{\textbf{Full supervision, \textsc{XGBGraph}.} Official GADBench
training masks (FullSup), test AUPRC ($\times100$), mean $\pm$ standard
deviation over ten seeds; not comparable with \Scarce{k}, whose test
nodes differ. Bold: best result in each row. Zero standard deviations:
repeated masks (see text).}
\label{tab:app-fullsup-xgb}

\begin{adjustbox}{max width=\linewidth}
\begin{tabular}{lcccccc}
\toprule
Dataset &
No resampling &
Class weighting &
ROS &
ROSE &
SMOTE-$\Phi$ &
Island view \\
\midrule

\multicolumn{7}{l}{\emph{Organic anomalies (real labels)}}\\
\midrule
Reddit & \(6.49 \pm 0.34\) & \(5.61 \pm 0.54\) & \(5.60 \pm 0.50\) & \(6.24 \pm 0.65\) & \(\mathbf{6.81 \pm 0.77}\) & \(6.08 \pm 0.66\) \\
Weibo & \(97.49 \pm 0.74\) & \(97.55 \pm 0.59\) & \(97.48 \pm 0.56\) & \(\mathbf{97.67 \pm 0.57}\) & \(97.05 \pm 0.73\) & \(97.63 \pm 0.35\) \\
Amazon & \(\mathbf{92.93 \pm 0.00}\) & \(92.61 \pm 0.00\) & \(92.70 \pm 0.26\) & \(92.89 \pm 0.34\) & \(92.25 \pm 0.41\) & \(92.62 \pm 0.37\) \\
YelpChi & \(\mathbf{88.29 \pm 0.00}\) & \(87.25 \pm 0.00\) & \(86.79 \pm 0.38\) & \(87.42 \pm 0.38\) & \(85.82 \pm 0.48\) & \(86.78 \pm 0.48\) \\
T-Finance & \(89.54 \pm 0.98\) & \(\mathbf{89.58 \pm 0.95}\) & \(89.54 \pm 0.97\) & \(88.96 \pm 0.93\) & \(89.11 \pm 1.06\) & \(88.64 \pm 0.95\) \\
Tolokers & \(49.53 \pm 1.44\) & \(49.06 \pm 1.22\) & \(48.57 \pm 2.22\) & \(\mathbf{49.68 \pm 1.75}\) & \(47.58 \pm 1.73\) & \(44.00 \pm 1.19\) \\
Questions & \(18.19 \pm 1.40\) & \(15.28 \pm 0.71\) & \(15.66 \pm 1.13\) & \(\mathbf{18.55 \pm 1.19}\) & \(17.10 \pm 0.90\) & \(15.49 \pm 0.93\) \\
FDCompCN & \(\mathbf{39.96 \pm 3.06}\) & \(39.01 \pm 2.29\) & \(39.18 \pm 2.46\) & \(39.63 \pm 2.95\) & \(37.68 \pm 3.01\) & \(38.85 \pm 2.17\) \\

\midrule
\multicolumn{7}{l}{\emph{Injected benchmarks (legacy)}}\\
\midrule
Cora & \(58.36 \pm 3.76\) & \(58.46 \pm 3.57\) & \(58.49 \pm 3.77\) & \(\mathbf{60.43 \pm 3.65}\) & \(59.54 \pm 3.26\) & \(59.76 \pm 4.39\) \\
Citeseer & \(62.80 \pm 6.29\) & \(61.90 \pm 5.98\) & \(61.93 \pm 5.89\) & \(65.43 \pm 6.03\) & \(63.51 \pm 5.57\) & \(\mathbf{65.51 \pm 3.91}\) \\
BlogCatalog & \(64.66 \pm 5.33\) & \(57.26 \pm 5.46\) & \(58.68 \pm 5.88\) & \(\mathbf{66.05 \pm 4.85}\) & \(59.35 \pm 5.72\) & \(58.60 \pm 5.21\) \\
Flickr & \(71.18 \pm 2.03\) & \(66.41 \pm 2.03\) & \(66.45 \pm 1.77\) & \(\mathbf{74.17 \pm 2.32}\) & \(70.50 \pm 2.52\) & \(62.23 \pm 3.09\) \\
ACM & \(56.04 \pm 2.28\) & \(54.34 \pm 2.62\) & \(54.67 \pm 1.77\) & \(\mathbf{58.37 \pm 2.39}\) & \(55.92 \pm 2.52\) & \(40.00 \pm 2.59\) \\
Pubmed & \(69.77 \pm 2.95\) & \(67.44 \pm 2.42\) & \(67.66 \pm 2.36\) & \(\mathbf{70.09 \pm 2.44}\) & \(66.84 \pm 2.66\) & \(68.16 \pm 2.92\) \\

\midrule
\multicolumn{7}{l}{\emph{Injected benchmarks (new supports)}}\\
\midrule
Amazon-Photo & \(72.65 \pm 2.65\) & \(\mathbf{73.66 \pm 3.93}\) & \(73.45 \pm 4.22\) & \(69.60 \pm 4.26\) & \(70.81 \pm 4.39\) & \(63.15 \pm 3.79\) \\
Amazon-Computers & \(\mathbf{67.72 \pm 1.91}\) & \(67.63 \pm 2.45\) & \(66.96 \pm 2.26\) & \(66.86 \pm 1.87\) & \(66.06 \pm 1.26\) & \(53.94 \pm 2.66\) \\
Coauthor-CS & \(\mathbf{93.39 \pm 1.07}\) & \(91.91 \pm 1.20\) & \(92.18 \pm 1.02\) & \(93.27 \pm 1.46\) & \(90.06 \pm 1.38\) & \(87.57 \pm 1.63\) \\

\midrule
\textbf{Macro average} & \(64.65\) & \(63.23\) & \(63.29\) & \(\mathbf{65.02}\) & \(63.29\) & \(60.53\) \\

\bottomrule
\end{tabular}
\end{adjustbox}
\end{table}

\begin{table}[!htbp]
\centering
\small
\setlength{\tabcolsep}{4pt}
\caption{\textbf{Full supervision, \textsc{RFGraph}.} Same layout as
Table~\ref{tab:app-fullsup-xgb}.}
\label{tab:app-fullsup-rf}

\begin{adjustbox}{max width=\linewidth}
\begin{tabular}{lcccccc}
\toprule
Dataset &
No resampling &
Class weighting &
ROS &
ROSE &
SMOTE-$\Phi$ &
Island view \\
\midrule

\multicolumn{7}{l}{\emph{Organic anomalies (real labels)}}\\
\midrule
Reddit & \(6.23 \pm 0.54\) & \(5.42 \pm 0.44\) & \(5.75 \pm 0.31\) & \(5.45 \pm 0.53\) & \(\mathbf{6.80 \pm 1.13}\) & \(5.55 \pm 0.34\) \\
Weibo & \(\mathbf{97.26 \pm 0.31}\) & \(96.98 \pm 0.27\) & \(97.12 \pm 0.35\) & \(96.67 \pm 0.41\) & \(96.86 \pm 0.40\) & \(97.21 \pm 0.43\) \\
Amazon & \(90.74 \pm 0.33\) & \(91.29 \pm 0.14\) & \(\mathbf{91.73 \pm 0.32}\) & \(91.30 \pm 0.18\) & \(91.08 \pm 0.42\) & \(89.79 \pm 0.33\) \\
YelpChi & \(76.57 \pm 0.44\) & \(76.57 \pm 0.30\) & \(\mathbf{77.73 \pm 0.35}\) & \(76.38 \pm 0.37\) & \(75.45 \pm 0.38\) & \(72.14 \pm 0.47\) \\
T-Finance & \(88.91 \pm 1.00\) & \(89.00 \pm 0.95\) & \(\mathbf{89.20 \pm 0.95}\) & \(87.42 \pm 1.04\) & \(88.98 \pm 0.99\) & \(87.19 \pm 1.01\) \\
Tolokers & \(\mathbf{51.07 \pm 1.99}\) & \(49.36 \pm 1.72\) & \(49.07 \pm 1.93\) & \(50.91 \pm 2.08\) & \(48.37 \pm 1.70\) & \(42.14 \pm 1.73\) \\
Questions & \(\mathbf{15.36 \pm 0.88}\) & \(14.23 \pm 1.25\) & \(12.87 \pm 1.29\) & \(14.05 \pm 0.95\) & \(15.35 \pm 1.25\) & \(12.75 \pm 1.21\) \\
FDCompCN & \(\mathbf{39.90 \pm 2.59}\) & \(38.99 \pm 2.36\) & \(39.06 \pm 3.17\) & \(39.57 \pm 2.75\) & \(35.33 \pm 2.45\) & \(38.56 \pm 2.53\) \\

\midrule
\multicolumn{7}{l}{\emph{Injected benchmarks (legacy)}}\\
\midrule
Cora & \(\mathbf{59.91 \pm 3.84}\) & \(58.00 \pm 3.48\) & \(57.28 \pm 3.70\) & \(59.84 \pm 3.80\) & \(59.69 \pm 3.86\) & \(59.05 \pm 4.19\) \\
Citeseer & \(62.12 \pm 4.02\) & \(60.35 \pm 5.07\) & \(61.06 \pm 3.83\) & \(61.91 \pm 4.49\) & \(61.72 \pm 4.90\) & \(\mathbf{63.94 \pm 3.57}\) \\
BlogCatalog & \(\mathbf{59.34 \pm 5.76}\) & \(43.27 \pm 7.51\) & \(51.28 \pm 7.80\) & \(57.43 \pm 6.30\) & \(50.44 \pm 5.66\) & \(55.05 \pm 5.89\) \\
Flickr & \(73.90 \pm 1.59\) & \(61.76 \pm 4.09\) & \(69.88 \pm 2.67\) & \(73.09 \pm 2.16\) & \(\mathbf{74.03 \pm 2.28}\) & \(68.21 \pm 2.13\) \\
ACM & \(65.12 \pm 2.87\) & \(54.21 \pm 2.35\) & \(54.85 \pm 2.11\) & \(\mathbf{65.83 \pm 2.51}\) & \(61.86 \pm 2.59\) & \(29.14 \pm 3.29\) \\
Pubmed & \(70.03 \pm 2.83\) & \(60.09 \pm 2.42\) & \(63.10 \pm 2.58\) & \(\mathbf{70.22 \pm 2.78}\) & \(61.47 \pm 1.82\) & \(64.47 \pm 2.56\) \\

\midrule
\multicolumn{7}{l}{\emph{Injected benchmarks (new supports)}}\\
\midrule
Amazon-Photo & \(53.85 \pm 3.94\) & \(69.18 \pm 4.04\) & \(\mathbf{71.34 \pm 3.44}\) & \(54.06 \pm 4.21\) & \(57.95 \pm 5.01\) & \(52.31 \pm 4.75\) \\
Amazon-Computers & \(48.07 \pm 2.39\) & \(55.73 \pm 2.70\) & \(\mathbf{59.10 \pm 2.94}\) & \(47.70 \pm 2.40\) & \(48.18 \pm 2.29\) & \(33.68 \pm 1.15\) \\
Coauthor-CS & \(\mathbf{92.10 \pm 1.29}\) & \(81.97 \pm 1.04\) & \(83.63 \pm 1.60\) & \(91.07 \pm 1.65\) & \(80.44 \pm 1.59\) & \(69.18 \pm 2.83\) \\

\midrule
\textbf{Macro average} & \(\mathbf{61.79}\) & \(59.20\) & \(60.83\) & \(61.35\) & \(59.65\) & \(55.31\) \\

\bottomrule
\end{tabular}
\end{adjustbox}
\end{table}

\FloatBarrier
\subsection{Cross-family tables}
\label{app:cross-family}
Tables~\ref{tab:app-cross-family-other} to~\ref{tab:app-ladder} complete
Table~\ref{tab:cross-family}: the three remaining samplers; the by-type
contrasts behind Figure~\ref{fig:rose}; the variation over splits and
strategies; the paired margins of the tabular detectors over six GNN
detectors; the effect of tuning; the two label-free columns; ROSE on
frozen GNN embeddings; both anomaly types under the two injection
recipes, then the same contrast on each graph; and the candidate ladder
on each graph.
\begin{widetables}
\begin{table}[!htbp]
\centering\small
\setlength{\tabcolsep}{3.2pt}
\caption{\textbf{Remaining samplers across detectors (\Scarce{20}).} Same columns as Table~\ref{tab:cross-family}. Views: island views; GODM: 169 of 170 trials.}
\label{tab:app-cross-family-other}
\begin{adjustbox}{max width=\linewidth}
\begin{tabular}{lrrrrrrrr}\toprule
& \multicolumn{4}{c}{Organic (8)} & \multicolumn{4}{c}{Injected (9)}\\
\cmidrule(lr){2-5}\cmidrule(lr){6-9}
Detector & Ref. & ADASYN & Views & GODM & Ref. & ADASYN & Views & GODM\\
\midrule
\multicolumn{9}{l}{\emph{Tabular foundation model, labelled rows in context (gains over the unresampled context)}}\\
TFM4GAD & $46.3$ & --- & --- & --- & $32.5$ & --- & --- & ---\\
TabPFN on $\Phi$ & $45.4$ & --- & $-0.88$ & $-0.51$ & $29.1$ & --- & $-1.51$ & $\mathbf{+1.26}$\\
\midrule
\multicolumn{9}{l}{\emph{Tree ensembles on $\Phi$ (gains over class weighting)}}\\
\textsc{RFGraph}, tuned & $44.8$ & --- & --- & --- & $29.4$ & --- & --- & ---\\
\textsc{XGBGraph}, tuned & $43.7$ & --- & --- & --- & $25.1$ & --- & --- & ---\\
\textsc{RFGraph}, default & $41.5$ & $+0.31$ & $-0.75$ & $-0.81$ & $25.8$ & $+1.21$ & $-2.93$ & $+1.64$\\
\textsc{XGBGraph}, default & $41.9$ & $-0.55$ & $-0.58$ & $-1.47$ & $20.7$ & $\mathbf{+3.25}$ & $-1.48$ & $\mathbf{+2.58}$\\
\midrule
\multicolumn{9}{l}{\emph{Graph neural networks, samplers on the hidden representation (gains over class weighting)}}\\
SpaceGNN & $40.1$ & $+0.01$ & --- & --- & $16.7$ & $+0.49$ & --- & ---\\
GHRN & $38.1$ & $-0.67$ & --- & --- & $22.4$ & $\mathbf{-1.01}$ & --- & ---\\
BWGNN & $37.9$ & $-0.22$ & $-0.42$ & $\mathbf{-1.70}$ & $22.8$ & $\mathbf{-1.19}$ & $-2.35$ & $\mathbf{-2.78}$\\
GATv2 & $23.0$ & $-1.05$ & $+0.71$ & $-1.21$ & $18.3$ & $-0.22$ & $-1.09$ & $-2.17$\\
\bottomrule\end{tabular}\end{adjustbox}
\end{table}

\begin{table}[!htbp]
\centering\small
\setlength{\tabcolsep}{4pt}
\caption{\textbf{ROSE by injected anomaly type, all detectors} (\Scarce{20}, eight injected benchmarks with type labels). Each type is scored against all normal test nodes. Ref.: AUPRC $\times100$ of the reference (class weighting; unweighted for the in-context models). $\Delta$: ROSE minus reference, 95\% interval, datasets improved. ROSE $-$ ROS: the same contrast against duplication, which isolates the noise.}
\label{tab:app-types-detectors}
\begin{adjustbox}{max width=\linewidth}
\begin{tabular}{lrllrll}\toprule
& \multicolumn{3}{c}{Attribute swaps (contextual)} & \multicolumn{3}{c}{Cliques (structural)}\\
\cmidrule(lr){2-4}\cmidrule(lr){5-7}
Detector & Ref. & $\Delta$ & ROSE $-$ ROS & Ref. & $\Delta$ & ROSE $-$ ROS\\\midrule
TFM4GAD & $22.3$ & $\mathbf{+2.64}$ $[+0.67,+5.01]$ (8/8) & $+2.08$ $[-2.33,+6.43]$ & $34.7$ & $\mathbf{-2.81}$ $[-6.03,-0.13]$ (1/8) & $\mathbf{-4.71}$ $[-9.29,-0.85]$\\
TabPFN on $\Phi$ & $11.3$ & $\mathbf{+3.60}$ $[+1.25,+6.42]$ (8/8) & $\mathbf{+6.38}$ $[+1.35,+12.45]$ & $39.3$ & $+0.25$ $[-1.30,+1.84]$ (4/8) & $-1.02$ $[-4.04,+1.52]$\\
\textsc{RFGraph}, tuned & $11.5$ & $\mathbf{+5.36}$ $[+0.88,+10.86]$ (5/8) & $\mathbf{+5.26}$ $[+1.26,+10.00]$ & $38.9$ & $-2.10$ $[-6.45,+2.68]$ (2/8) & $-2.72$ $[-6.57,+1.34]$\\
\textsc{XGBGraph}, tuned & $10.5$ & $\mathbf{+3.96}$ $[+0.65,+8.24]$ (7/8) & $\mathbf{+3.22}$ $[+0.57,+6.73]$ & $32.6$ & $+0.69$ $[-2.47,+4.38]$ (4/8) & $-0.03$ $[-3.48,+2.86]$\\
\textsc{RFGraph}, default & $5.2$ & $\mathbf{+10.77}$ $[+3.18,+19.65]$ (7/8) & $\mathbf{+10.68}$ $[+3.08,+19.59]$ & $39.9$ & $-3.54$ $[-8.96,+1.14]$ (3/8) & $-3.07$ $[-10.17,+2.93]$\\
\textsc{XGBGraph}, default & $6.4$ & $\mathbf{+5.64}$ $[+1.53,+10.81]$ (8/8) & $\mathbf{+5.75}$ $[+1.59,+10.93]$ & $29.2$ & $+3.89$ $[-1.73,+9.59]$ (5/8) & $+4.68$ $[-1.15,+10.63]$\\
SpaceGNN & $13.7$ & $+0.71$ $[-0.41,+1.89]$ (7/8) & $+0.27$ $[-0.78,+1.36]$ & $11.5$ & $+0.36$ $[-0.72,+1.70]$ (4/8) & $-0.68$ $[-2.80,+0.79]$\\
GHRN & $34.7$ & $-0.02$ $[-1.35,+1.19]$ (4/8) & $+0.63$ $[-1.01,+2.72]$ & $4.3$ & $+0.14$ $[-0.15,+0.56]$ (3/8) & $+0.01$ $[-0.21,+0.40]$\\
BWGNN & $35.3$ & $+0.11$ $[-1.38,+1.67]$ (4/8) & $+0.10$ $[-1.05,+1.21]$ & $4.6$ & $-0.04$ $[-0.46,+0.49]$ (2/8) & $-0.03$ $[-0.27,+0.23]$\\
GATv2 & $20.8$ & $-0.20$ $[-2.30,+1.88]$ (6/8) & $+0.71$ $[-1.26,+2.73]$ & $8.8$ & $-0.39$ $[-1.47,+0.76]$ (3/8) & $-0.28$ $[-1.16,+0.60]$\\
\bottomrule\end{tabular}\end{adjustbox}
\end{table}

\begin{table}[!htbp]
\centering\small
\caption{\textbf{Sources of organic AUPRC variation, all detectors.} Mean within-dataset shares (\%) across ten splits and the strategies available for each detector among its reference, unweighted training, ROS, ROSE and SMOTE (\Scarce{20}, eight organic datasets). In AUPRC points: standard deviation over splits of the strategy-averaged score, and range of the split-averaged score over strategies, both averaged over datasets.}
\label{tab:app-variance-detectors}
\begin{tabular}{lrrrrr}\toprule
& & \multicolumn{2}{c}{Share (\%)} & \multicolumn{2}{c}{AUPRC points}\\
\cmidrule(lr){3-4}\cmidrule(lr){5-6}
Detector & Strategies & Split & Strategy & SD over splits & Range over strategies\\\midrule
TFM4GAD & 3 & 79 & 10 & 2.8 & 2.0\\
TabPFN on $\Phi$ & 4 & 73 & 12 & 2.6 & 2.7\\
\textsc{RFGraph}, tuned & 5 & 59 & 9 & 2.5 & 2.4\\
\textsc{XGBGraph}, tuned & 5 & 64 & 7 & 2.7 & 2.1\\
\textsc{RFGraph}, default & 5 & 67 & 16 & 2.8 & 3.2\\
\textsc{XGBGraph}, default & 5 & 74 & 6 & 2.9 & 2.1\\
SpaceGNN & 5 & 79 & 3 & 2.2 & 1.1\\
GHRN & 5 & 53 & 11 & 3.1 & 4.0\\
BWGNN & 5 & 58 & 9 & 3.2 & 3.5\\
GATv2 & 5 & 60 & 3 & 3.8 & 1.9\\
\bottomrule\end{tabular}
\end{table}

\begin{table}[!htbp]
\centering\small
\setlength{\tabcolsep}{4pt}
\caption{\textbf{Paired margins over six GNN detectors} (\Scarce{20}, AUPRC points, 95\% interval). Class-weighted \textsc{RFGraph} at its default and tuned configuration, and TFM4GAD, minus each GNN at its published or default configuration. The GNNs were not tuned.}
\label{tab:app-margins}
\begin{adjustbox}{max width=\linewidth}
\begin{tabular}{lllllll}\toprule
& \multicolumn{2}{c}{\textsc{RFGraph}, default} & \multicolumn{2}{c}{\textsc{RFGraph}, tuned} & \multicolumn{2}{c}{TFM4GAD}\\
\cmidrule(lr){2-3}\cmidrule(lr){4-5}\cmidrule(lr){6-7}
Minus & Organic (8) & All 17 & Organic (8) & All 17 & Organic (8) & All 17\\\midrule
BWGNN & $+3.6$ $[-2.3,+9.9]$ & $+3.3$ $[-5.5,+12.7]$ & $\mathbf{+6.9}$ $[+2.2,+12.1]$ & $+6.7$ $[-1.5,+15.9]$ & $\mathbf{+8.4}$ $[+3.7,+13.2]$ & $\mathbf{+9.1}$ $[+2.7,+16.5]$\\
GraphSAGE & $\mathbf{+17.6}$ $[+4.9,+34.9]$ & $\mathbf{+11.0}$ $[+0.8,+22.3]$ & $\mathbf{+20.8}$ $[+6.6,+38.1]$ & $\mathbf{+14.4}$ $[+4.2,+25.7]$ & $\mathbf{+22.3}$ $[+7.7,+39.6]$ & $\mathbf{+16.7}$ $[+7.5,+27.1]$\\
GHRN & $+3.6$ $[-2.5,+9.7]$ & $+3.5$ $[-5.3,+12.9]$ & $\mathbf{+6.9}$ $[+2.4,+11.9]$ & $+6.9$ $[-1.2,+16.0]$ & $\mathbf{+8.4}$ $[+3.7,+13.2]$ & $\mathbf{+9.3}$ $[+2.9,+16.7]$\\
PC-GNN & $+6.6$ $[-1.3,+15.7]$ & $\mathbf{+8.5}$ $[+0.4,+17.6]$ & $\mathbf{+9.9}$ $[+3.5,+17.7]$ & $\mathbf{+11.9}$ $[+4.3,+20.7]$ & $\mathbf{+11.4}$ $[+4.6,+19.2]$ & $\mathbf{+14.3}$ $[+7.4,+22.0]$\\
SpaceGNN & $+1.4$ $[-3.3,+6.1]$ & $+5.4$ $[-0.1,+11.4]$ & $\mathbf{+4.7}$ $[+1.5,+8.8]$ & $\mathbf{+8.8}$ $[+3.9,+14.5]$ & $\mathbf{+6.2}$ $[+2.6,+10.0]$ & $\mathbf{+11.2}$ $[+6.5,+16.3]$\\
ConsisGAD & $+3.7$ $[-3.0,+11.1]$ & $\mathbf{+7.3}$ $[+0.7,+14.4]$ & $\mathbf{+7.0}$ $[+1.0,+14.9]$ & $\mathbf{+10.7}$ $[+4.6,+17.6]$ & $\mathbf{+8.5}$ $[+2.1,+17.2]$ & $\mathbf{+13.1}$ $[+7.6,+19.3]$\\
\bottomrule\end{tabular}\end{adjustbox}
\end{table}

\begin{table}[!htbp]
\centering\small
\setlength{\tabcolsep}{4pt}
\caption{\textbf{Tuning the tree detectors and ROSE} (\Scarce{20}, paired differences in AUPRC points, 95\% interval). CW: class weighting.}
\label{tab:app-tuning-effect}
\begin{adjustbox}{max width=\linewidth}
\begin{tabular}{lllll}\toprule
Detector & Contrast & Organic (8) & Injected (9) & All 17\\\midrule
\textsc{XGBGraph} & Tuned CW $-$ default CW & $\mathbf{+1.79}$ $[+0.16,+3.65]$ & $\mathbf{+4.31}$ $[+2.52,+6.14]$ & $\mathbf{+3.13}$ $[+1.77,+4.56]$\\
 & ROSE $-$ CW, default & $+0.94$ $[-0.33,+2.43]$ & $\mathbf{+5.22}$ $[+2.69,+7.44]$ & $\mathbf{+3.21}$ $[+1.49,+5.00]$\\
 & ROSE $-$ CW, tuned & $+0.91$ $[-0.03,+1.90]$ & $\mathbf{+2.57}$ $[+0.38,+4.89]$ & $\mathbf{+1.79}$ $[+0.54,+3.20]$\\
 & Change in ROSE $-$ CW after tuning & $-0.03$ $[-1.16,+0.95]$ & $\mathbf{-2.65}$ $[-4.89,-0.48]$ & $\mathbf{-1.42}$ $[-2.89,-0.09]$\\
\addlinespace[3pt]
\textsc{RFGraph} & Tuned CW $-$ default CW & $+3.26$ $[-0.01,+7.41]$ & $\mathbf{+3.51}$ $[+0.89,+6.16]$ & $\mathbf{+3.39}$ $[+1.28,+5.79]$\\
 & ROSE $-$ CW, default & $+0.99$ $[-1.08,+3.42]$ & $\mathbf{+4.40}$ $[+0.48,+8.53]$ & $\mathbf{+2.79}$ $[+0.42,+5.43]$\\
 & ROSE $-$ CW, tuned & $-0.52$ $[-1.76,+0.65]$ & $+1.64$ $[-1.26,+4.69]$ & $+0.62$ $[-1.03,+2.46]$\\
 & Change in ROSE $-$ CW after tuning & $-1.51$ $[-4.16,+0.55]$ & $-2.76$ $[-5.82,+0.41]$ & $\mathbf{-2.17}$ $[-4.27,-0.22]$\\
\bottomrule\end{tabular}\end{adjustbox}
\end{table}

\begin{table}[!htbp]
\centering\small
\setlength{\tabcolsep}{4pt}
\caption{\textbf{Two label-free columns appended to $\Phi$} (in-degree and $\lVert x\rVert_2$; default detectors, \Scarce{20}). Paired differences in AUPRC points with 95\% intervals. ROSE perturbs the two columns like any other. Swaps and cliques: the eight injected benchmarks with type labels.}
\label{tab:app-leak-columns}
\begin{adjustbox}{max width=\linewidth}
\begin{tabular}{llllll}\toprule
Detector & Contrast & Organic (8) & Injected (9) & Swaps (8) & Cliques (8)\\\midrule
\textsc{XGBGraph} & Class weighting: $\Phi$+2 minus $\Phi$ & $+0.71$ $[-0.12,+2.24]$ & $\mathbf{+8.11}$ $[+4.91,+11.67]$ & $\mathbf{+7.30}$ $[+5.08,+9.90]$ & $+1.25$ $[-0.45,+3.36]$\\
 & ROSE $-$ class weighting on $\Phi$ & $+0.94$ $[-0.33,+2.43]$ & $\mathbf{+5.22}$ $[+2.69,+7.44]$ & $\mathbf{+5.64}$ $[+1.53,+10.81]$ & $+3.89$ $[-1.73,+9.59]$\\
 & ROSE $-$ class weighting on $\Phi$+2 & $+0.90$ $[-0.54,+2.43]$ & $+2.46$ $[-0.43,+5.27]$ & $+4.26$ $[-0.34,+9.92]$ & $+1.60$ $[-2.86,+6.06]$\\
 & Change in ROSE $-$ class weighting & $-0.05$ $[-0.83,+0.65]$ & $\mathbf{-2.76}$ $[-4.99,-0.67]$ & $-1.38$ $[-3.46,+0.97]$ & $-2.30$ $[-5.55,+0.30]$\\
 & ROSE $-$ unweighted on $\Phi$+2 & $+0.83$ $[-0.36,+2.24]$ & $+0.28$ $[-2.27,+2.92]$ & --- & ---\\
\addlinespace[3pt]
\textsc{RFGraph} & Class weighting: $\Phi$+2 minus $\Phi$ & $+0.39$ $[-0.17,+1.18]$ & $\mathbf{+0.64}$ $[+0.01,+1.31]$ & $\mathbf{+0.38}$ $[+0.13,+0.62]$ & $+0.26$ $[-1.21,+1.76]$\\
 & ROSE $-$ class weighting on $\Phi$ & $+0.99$ $[-1.08,+3.42]$ & $\mathbf{+4.40}$ $[+0.48,+8.53]$ & $\mathbf{+10.77}$ $[+3.18,+19.65]$ & $-3.54$ $[-8.96,+1.14]$\\
 & ROSE $-$ class weighting on $\Phi$+2 & $+0.76$ $[-0.99,+2.90]$ & $\mathbf{+4.55}$ $[+0.85,+8.41]$ & $\mathbf{+11.18}$ $[+3.56,+20.11]$ & $-3.70$ $[-9.46,+1.01]$\\
 & Change in ROSE $-$ class weighting & $-0.22$ $[-1.04,+0.53]$ & $+0.15$ $[-0.70,+0.97]$ & $+0.41$ $[-0.20,+1.05]$ & $-0.15$ $[-1.83,+1.48]$\\
 & ROSE $-$ unweighted on $\Phi$+2 & $-0.09$ $[-1.48,+1.50]$ & $\mathbf{+3.40}$ $[+0.39,+6.87]$ & --- & ---\\
\bottomrule\end{tabular}\end{adjustbox}
\end{table}

\begin{table}[!htbp]
\centering\small
\caption{\textbf{ROSE on frozen GNN embeddings} (\Scarce{20}). A BWGNN or GraphSAGE encoder is trained on the split and frozen; its head input replaces $\Phi$ for a retrained head. ROSE minus class weighting, AUPRC points, 95\% interval. Ref.: AUPRC $\times100$ of the class-weighted head over 17 datasets.}
\label{tab:app-frozen}
\begin{tabular}{llrll}\toprule
Encoder & Head & Ref. & Organic (8) & Injected (9)\\\midrule
BWGNN & XGBoost & $24.2$ & $+0.32$ $[-0.31,+0.97]$ & $\mathbf{+1.49}$ $[+0.27,+2.73]$\\
 & Random forest & $28.5$ & $-0.16$ $[-0.68,+0.30]$ & $-0.71$ $[-1.84,+0.44]$\\
 & MLP & $29.9$ & $-0.02$ $[-0.98,+0.89]$ & $-0.22$ $[-0.73,+0.14]$\\
GraphSAGE & XGBoost & $16.6$ & $+0.37$ $[-0.16,+0.96]$ & $+0.57$ $[-0.43,+1.60]$\\
 & Random forest & $21.1$ & $+0.07$ $[-0.66,+0.69]$ & $-0.95$ $[-1.97,+0.07]$\\
 & MLP & $21.4$ & $-0.57$ $[-1.84,+0.19]$ & $+0.01$ $[-0.31,+0.32]$\\
\bottomrule\end{tabular}
\end{table}

\begin{table}[!htbp]
\centering\small
\setlength{\tabcolsep}{3.5pt}
\caption{\textbf{Two injection recipes on the same graphs, both anomaly types} (\Scarce{20}). Ref.: reference AUPRC $\times100$ on the type; $\Delta$: ROSE minus reference with its 95\% interval (graphs, realisations, splits). Norm lift and degree lift: AUPRC of the label-free score on attribute swaps (norm, with the more informative sign in each split) and on dense groups (degree), divided by prevalence.}
\label{tab:app-recipe}
\begin{adjustbox}{max width=\linewidth}
\begin{tabular}{lllrrrlrl}\toprule
& & & & & \multicolumn{2}{c}{Attribute swaps} & \multicolumn{2}{c}{Dense groups}\\
\cmidrule(lr){6-7}\cmidrule(lr){8-9}
Graphs & Detector & Recipe & Norm lift & Degree lift & Ref. & $\Delta$ & Ref. & $\Delta$\\
\midrule
Three bag-of-words graphs & \textsc{XGBGraph} & Standard & $17.3\times$ & $2.9\times$ & $13.4$ & $\mathbf{+11.61}$ $[+2.90,+18.98]$ & $8.4$ & $\mathbf{-2.04}$ $[-3.11,-1.13]$\\
 &  & Preserving & $1.1\times$ & $1.1\times$ & $2.6$ & $+0.13$ $[-0.02,+0.33]$ & $7.4$ & $\mathbf{-2.07}$ $[-2.67,-1.51]$\\
 & \textsc{RFGraph} & Standard & $17.3\times$ & $2.9\times$ & $11.5$ & $\mathbf{+19.79}$ $[+5.82,+30.76]$ & $16.1$ & $\mathbf{-8.93}$ $[-13.93,-4.37]$\\
 &  & Preserving & $1.1\times$ & $1.1\times$ & $2.6$ & $-0.03$ $[-0.11,+0.05]$ & $9.9$ & $\mathbf{-5.61}$ $[-7.97,-2.95]$\\
 & TabPFN on $\Phi$ & Standard & $17.3\times$ & $2.9\times$ & $23.5$ & $\mathbf{+5.79}$ $[+1.27,+9.89]$ & $9.0$ & $-0.75$ $[-2.71,+1.38]$\\
 &  & Preserving & $1.1\times$ & $1.1\times$ & $2.6$ & $+0.10$ $[-0.00,+0.21]$ & $8.0$ & $-0.95$ $[-3.59,+1.81]$\\
 & BWGNN & Standard & $17.3\times$ & $2.9\times$ & $55.8$ & $+2.37$ $[-1.64,+7.21]$ & $2.7$ & $-0.05$ $[-0.16,+0.06]$\\
 &  & Preserving & $1.1\times$ & $1.1\times$ & $3.1$ & $-0.07$ $[-0.31,+0.10]$ & $2.6$ & $+0.01$ $[-0.10,+0.12]$\\
\midrule
Six organic hosts & \textsc{XGBGraph} & Standard & $10.9\times$ & $9.5\times$ & $10.8$ & $\mathbf{+3.63}$ $[+1.14,+6.12]$ & $12.1$ & $+0.08$ $[-1.96,+2.41]$\\
 &  & Preserving & $1.1\times$ & $1.0\times$ & $4.4$ & $+0.56$ $[-0.08,+1.23]$ & $3.9$ & $-0.03$ $[-0.49,+0.42]$\\
 & \textsc{RFGraph} & Standard & $10.9\times$ & $9.5\times$ & $13.0$ & $\mathbf{+5.94}$ $[+1.41,+11.09]$ & $18.5$ & $\mathbf{-9.15}$ $[-16.71,-3.23]$\\
 &  & Preserving & $1.1\times$ & $1.0\times$ & $4.4$ & $+0.37$ $[-0.23,+0.96]$ & $4.5$ & $-0.90$ $[-1.99,+0.15]$\\
 & TabPFN on $\Phi$ & Standard & $10.9\times$ & $9.5\times$ & $17.6$ & $+2.39$ $[-2.79,+7.44]$ & $25.6$ & $\mathbf{-4.76}$ $[-8.63,-1.64]$\\
 &  & Preserving & $1.1\times$ & $1.0\times$ & $8.6$ & $-2.71$ $[-8.29,+0.26]$ & $7.1$ & $-1.94$ $[-4.89,+0.05]$\\
 & BWGNN & Standard & $10.9\times$ & $9.5\times$ & $18.3$ & $+0.34$ $[-0.75,+1.70]$ & $12.7$ & $+0.37$ $[-1.04,+2.54]$\\
 &  & Preserving & $1.1\times$ & $1.0\times$ & $4.4$ & $+0.05$ $[-0.24,+0.27]$ & $9.6$ & $-0.22$ $[-1.55,+1.09]$\\
\bottomrule\end{tabular}\end{adjustbox}
\end{table}

\begin{table}[!htbp]
\centering\small
\setlength{\tabcolsep}{3.5pt}
\caption{\textbf{Two injection recipes, per graph: attribute swaps} (\Scarce{20}, means over realisations and splits). S: standard recipe; P: norm- and degree-preserving. Ref.: reference AUPRC $\times100$ on the swaps; $\Delta$: ROSE minus reference.}
\label{tab:app-recipe-graphs}
\begin{adjustbox}{max width=\linewidth}
\begin{tabular}{lrrrrrrrrrrrrrrrrrr}\toprule
& \multicolumn{2}{c}{Norm lift} & \multicolumn{4}{c}{\textsc{XGBGraph}} & \multicolumn{4}{c}{\textsc{RFGraph}} & \multicolumn{4}{c}{TabPFN on $\Phi$} & \multicolumn{4}{c}{BWGNN}\\
\cmidrule(lr){2-3}\cmidrule(lr){4-7}\cmidrule(lr){8-11}\cmidrule(lr){12-15}\cmidrule(lr){16-19}
Graph & S & P & Ref. S & $\Delta$ S & Ref. P & $\Delta$ P & Ref. S & $\Delta$ S & Ref. P & $\Delta$ P & Ref. S & $\Delta$ S & Ref. P & $\Delta$ P & Ref. S & $\Delta$ S & Ref. P & $\Delta$ P\\\midrule
Amazon-Photo & $14.9\times$ & $1.1\times$ & $20.1$ & $+18.8$ & $2.5$ & $+0.3$ & $19.2$ & $+30.0$ & $2.5$ & $-0.1$ & $34.0$ & $+8.5$ & $2.5$ & $+0.2$ & $65.6$ & $+3.9$ & $3.1$ & $-0.2$\\
Amazon-Computers & $13.5\times$ & $1.1\times$ & $12.0$ & $+13.6$ & $2.6$ & $+0.1$ & $10.0$ & $+23.8$ & $2.6$ & $-0.0$ & $22.1$ & $+7.8$ & $2.6$ & $+0.1$ & $54.9$ & $+3.1$ & $3.1$ & $-0.1$\\
Coauthor-CS & $23.5\times$ & $1.1\times$ & $7.9$ & $+2.4$ & $2.7$ & $-0.0$ & $5.4$ & $+5.6$ & $2.6$ & $+0.0$ & $14.5$ & $+1.2$ & $2.6$ & $-0.0$ & $46.9$ & $+0.1$ & $3.2$ & $+0.0$\\
\addlinespace[3pt]
Weibo & $24.4\times$ & $1.2\times$ & $12.4$ & $+6.4$ & $3.2$ & $-0.2$ & $16.7$ & $+4.4$ & $3.7$ & $-0.6$ & $16.1$ & $+5.9$ & $4.6$ & $-0.4$ & $44.7$ & $+2.2$ & $2.4$ & $+0.0$\\
YelpChi & $1.2\times$ & $1.0\times$ & $3.6$ & $+1.2$ & $3.5$ & $+1.1$ & $3.3$ & $+1.5$ & $3.3$ & $+1.5$ & $4.1$ & $+1.0$ & $4.2$ & $+0.6$ & $4.6$ & $-0.1$ & $4.4$ & $+0.2$\\
Questions & $22.2\times$ & $1.0\times$ & $19.1$ & $+6.6$ & $3.3$ & $+0.5$ & $25.0$ & $+11.9$ & $3.7$ & $+0.7$ & $19.4$ & $+12.5$ & $3.6$ & $+0.4$ & $23.7$ & $+0.4$ & $3.8$ & $+0.2$\\
Reddit & $1.2\times$ & $1.1\times$ & $5.3$ & $-0.4$ & $3.9$ & $-0.4$ & $5.1$ & $-0.4$ & $3.6$ & $-0.3$ & $19.1$ & $-7.9$ & $22.6$ & $-15.2$ & $2.8$ & $-0.1$ & $2.6$ & $+0.0$\\
FDCompCN & $9.7\times$ & $1.2\times$ & $12.4$ & $+1.2$ & $6.5$ & $+1.7$ & $15.6$ & $+2.4$ & $6.5$ & $+0.4$ & $15.6$ & $-0.4$ & $8.2$ & $-1.2$ & $16.6$ & $-0.5$ & $8.2$ & $-0.4$\\
Tolokers & $6.9\times$ & $1.1\times$ & $12.4$ & $+6.8$ & $5.8$ & $+0.7$ & $12.4$ & $+15.8$ & $5.9$ & $+0.6$ & $31.3$ & $+3.2$ & $8.4$ & $-0.5$ & $17.5$ & $+0.1$ & $5.0$ & $+0.2$\\
\bottomrule\end{tabular}\end{adjustbox}
\end{table}

\begin{table}[!htbp]
\centering\small
\setlength{\tabcolsep}{3pt}
\caption{\textbf{Candidate ladder, per graph: attribute swaps} (\Scarce{20}, standard recipe with $k$ candidates per swap, means over two realisations and ten splits). The realisations of a graph share their dense groups and swap targets across $k$. $\Delta$: ROSE minus class weighting on the swaps; Ref.: class-weighted AUPRC $\times100$ on the swaps (prevalence about 2.4\%). BWGNN gains at most $+3.9$ on any graph and $k$.}
\label{tab:app-ladder}
\begin{adjustbox}{max width=\linewidth}
\begin{tabular}{lrrrrrrrrrrrrrrrrrrrr}\toprule
 & \multicolumn{5}{c}{Norm lift} & \multicolumn{5}{c}{\textsc{XGBGraph}, $\Delta$} & \multicolumn{5}{c}{\textsc{RFGraph}, $\Delta$} & \multicolumn{5}{c}{BWGNN, Ref.}\\
\cmidrule(lr){2-6}\cmidrule(lr){7-11}\cmidrule(lr){12-16}\cmidrule(lr){17-21}
Graph, $k=$ & 1 & 2 & 5 & 15 & 50 & 1 & 2 & 5 & 15 & 50 & 1 & 2 & 5 & 15 & 50 & 1 & 2 & 5 & 15 & 50\\\midrule
Amazon-Photo & $1.1\times$ & $1.3\times$ & $2.6\times$ & $7.1\times$ & $14.9\times$ & $+0.1$ & $+0.6$ & $+3.6$ & $+10.8$ & $+18.8$ & $+0.1$ & $+1.0$ & $+7.3$ & $+20.8$ & $+30.0$ & $4.6$ & $8.2$ & $24.7$ & $54.0$ & $65.6$\\
Amazon-Computers & $1.1\times$ & $1.3\times$ & $2.3\times$ & $5.6\times$ & $13.5\times$ & $+0.1$ & $+0.8$ & $+1.1$ & $+5.9$ & $+13.6$ & $+0.0$ & $+1.1$ & $+3.9$ & $+13.1$ & $+23.8$ & $3.5$ & $5.7$ & $14.7$ & $38.8$ & $54.9$\\
Coauthor-CS & $1.0\times$ & $1.6\times$ & $3.3\times$ & $9.3\times$ & $23.5\times$ & $-0.0$ & $-0.0$ & $+0.1$ & $+0.7$ & $+2.4$ & $-0.0$ & $+0.0$ & $+0.1$ & $+0.9$ & $+5.6$ & $3.7$ & $3.3$ & $5.0$ & $14.7$ & $46.9$\\
\addlinespace[3pt]
Weibo & $1.0\times$ & $1.7\times$ & $3.7\times$ & $9.7\times$ & $24.9\times$ & $-0.5$ & $-0.6$ & $-1.0$ & $-0.5$ & $+7.2$ & $-1.0$ & $-1.3$ & $-5.1$ & $-3.5$ & $+3.7$ & $2.3$ & $3.0$ & $6.4$ & $15.1$ & $43.9$\\
YelpChi & $1.0\times$ & $1.0\times$ & $1.1\times$ & $1.1\times$ & $1.3\times$ & $+0.0$ & $+0.1$ & $+0.3$ & $+0.8$ & $+1.1$ & $+0.0$ & $+0.1$ & $+0.4$ & $+0.8$ & $+1.5$ & $2.6$ & $2.7$ & $3.0$ & $3.8$ & $4.5$\\
Questions & $1.0\times$ & $1.5\times$ & $3.0\times$ & $7.9\times$ & $22.1\times$ & $+0.0$ & $+0.1$ & $+0.4$ & $+2.3$ & $+8.0$ & $-0.1$ & $-0.1$ & $+0.4$ & $+5.7$ & $+12.4$ & $2.5$ & $2.6$ & $3.3$ & $5.0$ & $24.8$\\
Reddit & $1.0\times$ & $1.2\times$ & $1.2\times$ & $1.2\times$ & $1.1\times$ & $-0.2$ & $-0.6$ & $-0.1$ & $+0.2$ & $-0.4$ & $-0.1$ & $-0.8$ & $-0.5$ & $-0.1$ & $-0.3$ & $2.5$ & $2.6$ & $2.6$ & $2.6$ & $2.8$\\
FDCompCN & $1.1\times$ & $1.3\times$ & $1.8\times$ & $3.3\times$ & $8.6\times$ & $-0.1$ & $-0.0$ & $+0.0$ & $+0.0$ & $+1.2$ & $-0.2$ & $-0.0$ & $+0.1$ & $+0.3$ & $+2.0$ & $2.7$ & $2.9$ & $3.9$ & $6.3$ & $16.7$\\
Tolokers & $1.1\times$ & $1.3\times$ & $1.7\times$ & $3.0\times$ & $6.9\times$ & $+0.1$ & $+0.4$ & $+0.7$ & $+2.5$ & $+7.4$ & $+0.1$ & $+0.4$ & $+1.1$ & $+5.0$ & $+15.3$ & $2.5$ & $2.9$ & $3.8$ & $7.5$ & $17.2$\\
\bottomrule\end{tabular}\end{adjustbox}
\end{table}

\end{widetables}

\FloatBarrier
\subsection{Anchors to published numbers}
\label{app:anchors}
We checked our implementations against published numbers on the seven
GADBench datasets of our organic panel; no published number exists at this
budget for the nine injected datasets. On GADBench's official
semi-supervised masks, the default trees reproduce GADBench's Table~11
\citep{tang2023gadbenchrevisitingbenchmarkingsupervised} within 0.15 AUPRC
points on every dataset. On our \Scarce{20} splits, whose draws differ,
the mean AUPRC of BWGNN, GHRN, GraphSAGE \citep{inductive}, SpaceGNN and ConsisGAD is within
0.6 points of the published mean at the same budget, that of PC-GNN within
1.2, and that of TFM4GAD within 0.9 on the three datasets its paper shares
with our panel. For SpaceGNN and ConsisGAD the published values are
re-runs reported by \citet{apf}; SpaceGNN's own random-search results give
a mean of 48.1, 4.0 points above our run. The table is in
\texttt{IMPLEMENTATION.md} of the released repository.

\FloatBarrier
\subsection{Diagnostic experiments}
\label{app:revision-tables}
All experiments in this subsection are diagnostic
(Appendix~\ref{sec:protocol}): the label-free scorers
(Table~\ref{tab:app-labelfree}), the same-graph injection and its
degree-adapted sensitivity (Table~\ref{tab:app-samegraph}), the adaptive
controls with \textsc{RFGraph} (Table~\ref{tab:app-rf-adaptive}) and ROSE
restricted to one feature block (Table~\ref{tab:app-blocks}). AUROC and
Recall@$K$ are given beside AUPRC for each primary contrast. Per-dataset
values of the last two tables and ROSE's gain by anomaly type on each
injected benchmark are in the released repository.

\begin{widetables}
\begin{table}[!htbp]
\centering
\small
\setlength{\tabcolsep}{3pt}
\caption{\textbf{Label-free scorers} (no training, no labels; test nodes of the ten \Scarce{20} splits,
mean over splits, $\times100$). \emph{All anomalies}: AUPRC of degree, $\lVert x\rVert$ and
$-\lVert x\rVert$; Prev.: test prevalence, the AUPRC of a random score; Best: AUROC and Rec@$K$ of
the scorer with the highest AUPRC on that dataset. Last two blocks: $\lVert x\rVert$ on contextual and
degree on structural anomalies, each against all test normals, on the eight injected datasets with
type labels.}
\label{tab:app-labelfree}
\begin{adjustbox}{max width=\linewidth}
\begin{tabular}{lrrrrlrrcrrrrrr}
\toprule
& \multicolumn{7}{c}{All anomalies} & & \multicolumn{3}{c}{Contextual: $\lVert x\rVert$} & \multicolumn{3}{c}{Structural: degree} \\
\cmidrule(lr){2-8}\cmidrule(lr){10-12}\cmidrule(lr){13-15}
& & \multicolumn{3}{c}{AUPRC} & \multicolumn{3}{c}{Best} & & & & & & & \\
\cmidrule(lr){3-5}\cmidrule(lr){6-8}
Dataset & Prev. & Deg. & $\lVert x\rVert$ & $-\lVert x\rVert$ & Scorer & AUROC & Rec@$K$ & & Prev. & AUPRC & AUROC & Prev. & AUPRC & AUROC \\
\midrule
Amazon & 8.9 & 5.3 & 7.1 & \textbf{9.3} & $-\lVert x\rVert$ & 56.6 & 2.8 &  & --- & --- & --- & --- & --- & --- \\
FDCompCN & 9.6 & 9.2 & 9.7 & \textbf{10.6} & $-\lVert x\rVert$ & 50.9 & 13.0 &  & --- & --- & --- & --- & --- & --- \\
Questions & 2.9 & \textbf{5.3} & 3.6 & 3.0 & Deg. & 61.1 & 9.9 &  & --- & --- & --- & --- & --- & --- \\
Reddit & 2.8 & \textbf{3.1} & 3.0 & 3.0 & Deg. & 55.8 & 0.8 &  & --- & --- & --- & --- & --- & --- \\
T-Finance & 4.4 & 5.2 & 2.4 & \textbf{29.9} & $-\lVert x\rVert$ & 85.3 & 35.9 &  & --- & --- & --- & --- & --- & --- \\
Tolokers & 21.7 & \textbf{28.5} & 21.4 & 21.5 & Deg. & 55.7 & 28.0 &  & --- & --- & --- & --- & --- & --- \\
Weibo & 9.8 & 10.6 & \textbf{80.4} & 5.2 & $\lVert x\rVert$ & 92.5 & 77.0 &  & --- & --- & --- & --- & --- & --- \\
YelpChi & 14.4 & 13.9 & 12.2 & \textbf{17.1} & $-\lVert x\rVert$ & 56.5 & 18.9 &  & --- & --- & --- & --- & --- & --- \\
\midrule
ACM & 3.2 & 9.1 & \textbf{25.5} & 2.0 & $\lVert x\rVert$ & 74.8 & 37.0 &  & --- & --- & --- & --- & --- & --- \\
BlogCatalog & 4.6 & 5.5 & \textbf{33.1} & 3.2 & $\lVert x\rVert$ & 74.6 & 44.5 &  & 2.4 & 55.8 & 98.7 & 2.4 & 3.2 & 61.2 \\
Citeseer & 2.6 & \textbf{24.1} & 16.3 & 1.8 & Deg. & 73.0 & 49.4 &  & 1.3 & 26.6 & 97.2 & 1.3 & 43.3 & 99.2 \\
Cora & 3.2 & \textbf{23.5} & 15.7 & 2.2 & Deg. & 76.2 & 44.9 &  & 1.6 & 23.0 & 95.9 & 1.7 & 36.9 & 98.6 \\
Flickr & 5.2 & 5.8 & \textbf{37.6} & 3.3 & $\lVert x\rVert$ & 74.3 & 44.8 &  & 2.6 & 63.8 & 98.7 & 2.7 & 3.5 & 64.6 \\
Pubmed & 2.7 & 8.7 & \textbf{21.6} & 1.8 & $\lVert x\rVert$ & 74.4 & 33.2 &  & 1.4 & 35.1 & 98.0 & 1.4 & 11.4 & 94.3 \\
Amazon-Computers & 4.6 & 5.6 & \textbf{21.6} & 5.5 & $\lVert x\rVert$ & 62.9 & 33.7 &  & 2.4 & 31.6 & 74.2 & 2.3 & 3.5 & 69.1 \\
Amazon-Photo & 4.2 & 5.4 & \textbf{24.5} & 6.5 & $\lVert x\rVert$ & 62.4 & 34.5 &  & 2.3 & 36.0 & 72.0 & 2.0 & 3.5 & 70.1 \\
Coauthor-CS & 4.7 & 11.3 & \textbf{34.6} & 3.0 & $\lVert x\rVert$ & 75.1 & 44.6 &  & 2.4 & 56.5 & 98.6 & 2.3 & 12.2 & 91.8 \\
\bottomrule
\end{tabular}
\end{adjustbox}
\end{table}

\begin{table}[!htbp]
\centering
\small
\setlength{\tabcolsep}{4pt}
\caption{\textbf{Same injection recipe on organic graphs, per host.} Paired ROSE $-$ class weighting on the injected task (DOMINANT/CoLA recipe unchanged; three realisations $\times$ ten splits) and on the organic task of the same graph (\Scarce{20}, $\times100$). \emph{Leak}: AUPRC of the better of $\lVert x\rVert$ and $-\lVert x\rVert$ on contextual anomalies, divided by their prevalence. \emph{All}: AUPRC, then AUROC and Recall@$K$; the type columns are AUPRC against all test normals. \emph{Deg.-adapted}: \emph{All} on one realisation with the clique size set from the host degree, without intervals. Macro: mean of the five primary hosts with its 95\% interval, bold when it excludes zero. Tolokers$^\dagger$ is an extra host, outside the macro. Hosts keep their native edge directions, unlike in Tables~\ref{tab:recipe} and~\ref{tab:app-recipe-graphs}.}
\label{tab:app-samegraph}
\begin{adjustbox}{max width=\linewidth}
\begin{tabular}{lrrrrrrrrrr}
\toprule
& & & \multicolumn{6}{c}{Injected task} & Organic task & \\
\cmidrule(lr){4-9}\cmidrule(lr){10-10}
Host & Dim. & Leak & All & AUROC & Rec@$K$ & Contextual & Structural & Deg.-adapted & All & Inj.\ $-$ org. \\
\midrule
\multicolumn{11}{l}{\emph{\textsc{XGBGraph}}}\\
\midrule
Weibo & 400 & 24.4$\times$ & +1.55 & +1.19 & +2.55 & +6.57 & $-$4.72 & +5.50 & +3.61 & $-$2.06 \\
Questions & 301 & 22.2$\times$ & +8.68 & +8.53 & +7.67 & +6.30 & +5.11 & +2.80 & +0.28 & +8.40 \\
FDCompCN & 57 & 9.7$\times$ & +0.57 & +2.54 & +3.31 & +1.21 & $-$2.74 & +1.90 & +1.40 & $-$0.83 \\
YelpChi & 32 & 1.2$\times$ & +0.83 & +2.72 & +1.86 & +1.22 & $-$0.04 & +0.65 & +4.14 & $-$3.31 \\
Reddit & 64 & 1.2$\times$ & +0.93 & +0.78 & +1.84 & $-$0.40 & +2.23 & +1.61 & $-$0.06 & +0.99 \\
\addlinespace
Macro (5 hosts) & & & \textbf{+2.51} & \textbf{+3.15} & \textbf{+3.45} & \textbf{+2.98} & $-$0.03 & +2.49 & \textbf{+1.87} & +0.64 \\
 & & & {\scriptsize[+0.29, +5.85]} & {\scriptsize[+1.17, +6.03]} & {\scriptsize[+1.63, +5.91]} & {\scriptsize[+0.47, +5.67]} & {\scriptsize[$-$3.35, +3.44]} &  & {\scriptsize[+0.30, +3.71]} & {\scriptsize[$-$2.66, +4.87]} \\
\addlinespace
Tolokers$^\dagger$ & 10 & 6.9$\times$ & +0.48 & +0.95 & +2.14 & +4.62 & $-$5.17 & --- & $-$1.18 & +1.67 \\
\midrule
\multicolumn{11}{l}{\emph{\textsc{RFGraph}}}\\
\midrule
Weibo & 400 & 24.4$\times$ & $-$3.69 & $-$3.61 & $-$1.13 & +8.80 & $-$15.04 & +0.36 & +6.99 & $-$10.68 \\
Questions & 301 & 22.2$\times$ & +0.37 & $-$7.47 & +3.61 & +13.74 & $-$6.92 & +2.94 & $-$1.40 & +1.77 \\
FDCompCN & 57 & 9.7$\times$ & $-$8.92 & +0.94 & $-$5.42 & +2.38 & $-$25.50 & +1.12 & +0.87 & $-$9.79 \\
YelpChi & 32 & 1.2$\times$ & +0.54 & +1.16 & +1.46 & +1.47 & $-$0.47 & $-$0.51 & +4.96 & $-$4.41 \\
Reddit & 64 & 1.2$\times$ & $-$6.95 & $-$4.45 & $-$7.22 & $-$0.38 & $-$12.14 & $-$1.80 & $-$0.24 & $-$6.71 \\
\addlinespace
Macro (5 hosts) & & & \textbf{$-$3.73} & $-$2.69 & $-$1.74 & \textbf{+5.20} & \textbf{$-$12.02} & +0.42 & +2.24 & \textbf{$-$5.96} \\
 & & & {\scriptsize[$-$7.26, $-$0.26]} & {\scriptsize[$-$5.64, +0.17]} & {\scriptsize[$-$5.59, +1.93]} & {\scriptsize[+0.91, +10.16]} & {\scriptsize[$-$19.97, $-$4.90]} &  & {\scriptsize[$-$0.51, +5.41]} & {\scriptsize[$-$10.00, $-$1.66]} \\
\addlinespace
Tolokers$^\dagger$ & 10 & 6.9$\times$ & $-$1.88 & +0.38 & $-$0.15 & +10.20 & $-$16.11 & --- & $-$1.46 & $-$0.42 \\
\bottomrule
\end{tabular}
\end{adjustbox}
\end{table}

\begin{table}[!htbp]
\centering
\small
\setlength{\tabcolsep}{3pt}
\caption{\textbf{Adaptive A/T/M controls with \textsc{RFGraph}}
(\Scarce{20}, paired $\Delta\times100$; 95\% intervals below, bold when
they exclude zero). Sources and generated rows are those of the XGBoost
mechanism controls. A default \textsc{RFGraph} is fitted on them as one
forest (\emph{pooled}, primary) or with its trees following the booster's
block schedule (\emph{blockwise}). XGB: the booster of the same trial (200
trees, learning rate 0.05). Lower panel: the other contrasts of
Table~\ref{tab:matched}, in AUPRC.}
\label{tab:app-rf-adaptive}
\begin{adjustbox}{max width=\linewidth}
\begin{tabular}{lrrrrrrrrrr}
\toprule
& \multicolumn{7}{c}{Views $-$ copies, same sources and mass (primary)}
& \multicolumn{3}{c}{Views (fixed mass) $-$ CW} \\
\cmidrule(lr){2-8}\cmidrule(lr){9-11}
& \multicolumn{3}{c}{RF pooled} & \multicolumn{3}{c}{RF blockwise} & XGB
& RF pooled & RF blockwise & XGB \\
\cmidrule(lr){2-4}\cmidrule(lr){5-7}\cmidrule(lr){8-8}\cmidrule(lr){9-9}\cmidrule(lr){10-10}\cmidrule(lr){11-11}
Panel & AUPRC & AUROC & Rec@$K$ & AUPRC & AUROC & Rec@$K$ & AUPRC & AUPRC & AUPRC & AUPRC \\
\midrule
\textbf{Organic (8)} & $+0.87$ & $+0.36$ & $+0.75$ & $+0.43$ & $+0.23$ & $+0.42$ & $+0.28$ & $\mathbf{+1.40}$ & $\mathbf{+0.60}$ & $+0.48$ \\
 & {\scriptsize$[-0.34, +2.41]$} & {\scriptsize$[-0.16, +0.88]$} & {\scriptsize$[-0.38, +2.06]$} & {\scriptsize$[-0.05, +1.08]$} & {\scriptsize$[0.00, +0.47]$} & {\scriptsize$[-0.12, +1.11]$} & {\scriptsize$[-0.18, +0.94]$} & {\scriptsize$[+0.29, +2.80]$} & {\scriptsize$[+0.07, +1.32]$} & {\scriptsize$[-0.08, +1.27]$} \\[2pt]
\textbf{Injected (9)} & $+0.15$ & $+0.37$ & $-0.09$ & $+0.46$ & $+0.30$ & $+0.20$ & $-0.32$ & $+0.36$ & $+0.50$ & $+0.01$ \\
 & {\scriptsize$[-0.69, +0.90]$} & {\scriptsize$[-0.66, +1.45]$} & {\scriptsize$[-0.83, +0.64]$} & {\scriptsize$[-0.06, +1.03]$} & {\scriptsize$[-0.19, +0.81]$} & {\scriptsize$[-0.43, +0.87]$} & {\scriptsize$[-1.40, +0.59]$} & {\scriptsize$[-0.58, +1.30]$} & {\scriptsize$[-0.02, +1.07]$} & {\scriptsize$[-1.21, +0.99]$} \\[2pt]
\textbf{All (17)} & $+0.49$ & $+0.36$ & $+0.31$ & $\mathbf{+0.44}$ & $+0.27$ & $+0.30$ & $-0.04$ & $\mathbf{+0.85}$ & $\mathbf{+0.55}$ & $+0.23$ \\
 & {\scriptsize$[-0.26, +1.34]$} & {\scriptsize$[-0.23, +0.99]$} & {\scriptsize$[-0.38, +1.08]$} & {\scriptsize$[+0.07, +0.87]$} & {\scriptsize$[-0.01, +0.56]$} & {\scriptsize$[-0.13, +0.78]$} & {\scriptsize$[-0.69, +0.53]$} & {\scriptsize$[+0.07, +1.71]$} & {\scriptsize$[+0.16, +0.99]$} & {\scriptsize$[-0.50, +0.87]$} \\[2pt]
\bottomrule
\end{tabular}
\end{adjustbox}

\vspace{6pt}
\begin{adjustbox}{max width=\linewidth}
\begin{tabular}{llccc}
\toprule
Contrast (AUPRC) & Panel & RF pooled & RF blockwise & XGB \\
\midrule
Added $-$ fixed mass (views) & Organic (8) & $\mathbf{-1.26}$ {\scriptsize$[-2.92, -0.13]$} & $-0.18$ {\scriptsize$[-0.50, +0.11]$} & $+0.02$ {\scriptsize$[-0.29, +0.40]$} \\
 & Injected (9) & $\mathbf{-2.26}$ {\scriptsize$[-4.73, -0.32]$} & $-0.33$ {\scriptsize$[-0.85, +0.17]$} & $-0.09$ {\scriptsize$[-0.68, +0.50]$} \\
SMOTE-$\Phi$ $-$ copies & Organic (8) & $-0.21$ {\scriptsize$[-0.86, +0.35]$} & $-0.08$ {\scriptsize$[-0.46, +0.23]$} & $+0.23$ {\scriptsize$[-0.12, +0.63]$} \\
 & Injected (9) & $\mathbf{+1.68}$ {\scriptsize$[+0.89, +2.49]$} & $+0.49$ {\scriptsize$[+0.00, +1.09]$} & $\mathbf{+0.96}$ {\scriptsize$[+0.45, +1.49]$} \\
Residual weights $-$ CW & Organic (8) & $+0.23$ {\scriptsize$[-0.12, +0.71]$} & $+0.24$ {\scriptsize$[-0.10, +0.76]$} & $\mathbf{+0.20}$ {\scriptsize$[+0.02, +0.41]$} \\
 & Injected (9) & $+0.04$ {\scriptsize$[-0.42, +0.57]$} & $-0.10$ {\scriptsize$[-0.48, +0.30]$} & $\mathbf{+0.33}$ {\scriptsize$[+0.05, +0.59]$} \\
SMOTE-$\Phi$ $-$ CW & Organic (8) & $+0.32$ {\scriptsize$[-0.47, +1.24]$} & $+0.09$ {\scriptsize$[-0.25, +0.46]$} & $\mathbf{+0.43}$ {\scriptsize$[+0.08, +0.83]$} \\
 & Injected (9) & $\mathbf{+1.89}$ {\scriptsize$[+0.89, +2.95]$} & $\mathbf{+0.53}$ {\scriptsize$[+0.07, +1.13]$} & $\mathbf{+1.29}$ {\scriptsize$[+0.76, +1.86]$} \\
\bottomrule
\end{tabular}
\end{adjustbox}
\end{table}

\begin{table}[!htbp]
\centering
\small
\setlength{\tabcolsep}{3pt}
\caption{\textbf{ROSE restricted to one feature block}
(\Scarce{20}, paired $\Delta$ against class weighting, $\times100$; 95\%
intervals below, bold when they exclude zero). Synthetic rows are
perturbed only in the columns of $X$, of $PX$, or of $P^2X$; the other
blocks keep the source row's values. Sources and noise are those of full
ROSE. AUROC and Recall@$K$ are given for the arm we designated as primary
before running this experiment, ROSE on $X$ only.}
\label{tab:app-blocks}
\begin{adjustbox}{max width=\linewidth}
\begin{tabular}{lrrrrrrrrrrrr}
\toprule
& \multicolumn{8}{c}{AUPRC} & \multicolumn{4}{c}{ROSE on $X$ only} \\
\cmidrule(lr){2-9}\cmidrule(lr){10-13}
& \multicolumn{4}{c}{\textsc{XGBGraph}} & \multicolumn{4}{c}{\textsc{RFGraph}}
& \multicolumn{2}{c}{\textsc{XGBGraph}} & \multicolumn{2}{c}{\textsc{RFGraph}} \\
\cmidrule(lr){2-5}\cmidrule(lr){6-9}\cmidrule(lr){10-11}\cmidrule(lr){12-13}
Panel & $X$ & $PX$ & $P^2X$ & Full & $X$ & $PX$ & $P^2X$ & Full
& AUROC & Rec@$K$ & AUROC & Rec@$K$ \\
\midrule
\textbf{Organic (8)} & $-0.22$ & $+0.48$ & $-0.43$ & $+0.94$ & $+0.06$ & $\mathbf{+1.38}$ & $+1.06$ & $+0.99$ & $+0.16$ & $-0.08$ & $+0.42$ & $+0.62$ \\
 & {\scriptsize$[-1.33, +0.86]$} & {\scriptsize$[-0.01, +1.13]$} & {\scriptsize$[-1.54, +0.35]$} & {\scriptsize$[-0.34, +2.45]$} & {\scriptsize$[-1.91, +2.08]$} & {\scriptsize$[+0.31, +2.66]$} & {\scriptsize$[-0.05, +2.30]$} & {\scriptsize$[-1.06, +3.47]$} & {\scriptsize$[-0.46, +0.83]$} & {\scriptsize$[-1.05, +0.92]$} & {\scriptsize$[-0.53, +1.60]$} & {\scriptsize$[-1.00, +2.46]$} \\[2pt]
\textbf{Injected (9)} & $\mathbf{+2.58}$ & $\mathbf{+2.60}$ & $+1.36$ & $\mathbf{+5.22}$ & $\mathbf{+4.14}$ & $+1.62$ & $-0.37$ & $\mathbf{+4.40}$ & $\mathbf{+2.01}$ & $\mathbf{+2.96}$ & $+3.05$ & $\mathbf{+4.00}$ \\
 & {\scriptsize$[+1.07, +4.33]$} & {\scriptsize$[+0.35, +4.91]$} & {\scriptsize$[-0.34, +3.11]$} & {\scriptsize$[+2.68, +7.43]$} & {\scriptsize$[+1.24, +7.77]$} & {\scriptsize$[-1.14, +3.92]$} & {\scriptsize$[-2.13, +1.19]$} & {\scriptsize$[+0.48, +8.55]$} & {\scriptsize$[+0.66, +3.65]$} & {\scriptsize$[+1.01, +5.27]$} & {\scriptsize$[-0.45, +6.86]$} & {\scriptsize$[+0.51, +8.47]$} \\[2pt]
\bottomrule
\end{tabular}
\end{adjustbox}
\end{table}
\end{widetables}

\end{document}